\documentclass{article}

\PassOptionsToPackage{numbers, compress}{natbib}
 \usepackage[preprint]{neurips_2026}

\usepackage[utf8]{inputenc} % allow utf-8 input
\usepackage[T1]{fontenc}    % use 8-bit T1 fonts
\usepackage{hyperref}       % hyperlinks
\usepackage{url}            % simple URL typesetting
\usepackage{booktabs}       % professional-quality tables
\usepackage{amsfonts}       % blackboard math symbols
\usepackage{nicefrac}       % compact symbols for 1/2, etc.
\usepackage{microtype}      % microtypography
\usepackage{xcolor}         % colors

\usepackage{graphicx}
\usepackage{subcaption}
\usepackage{amsmath}
\usepackage{hyperref}
\usepackage{algorithm}
\usepackage{algorithmic}
\usepackage{tabularx}
\usepackage{wrapfig}

\title{Adaptive Control in Autonomous Driving \\ via Real-Time Recurrent RL}

\author{%
  Julian Lemmel\textsuperscript{\rm 1*}\\
  \And
  Felix Resch\textsuperscript{\rm 1*} \\
  \And
  Mónika Farsang\textsuperscript{\rm 1} \\
  \AND
  Ramin Hasani\textsuperscript{\rm 2,3} \\
  \And
  Daniela Rus\textsuperscript{\rm 2,3} \\
  \And
  Radu Grosu\textsuperscript{\rm 1} \\
}

\begin{document}

\maketitle
\let\thefootnote\relax\footnotetext{\textsuperscript{\rm *}Equal contribution}
\let\thefootnote\relax\footnotetext{\textsuperscript{\rm 1}TU Wien}
\let\thefootnote\relax\footnotetext{\textsuperscript{\rm 2}MIT CSAIL}
\let\thefootnote\relax\footnotetext{\textsuperscript{\rm 3}Liquid AI}
\let\thefootnote\relax\footnotetext{\textsuperscript{\rm }Corr. author: julian.lemmel@tuwien.ac.at}

\begin{abstract}
We study online fine-tuning of pretrained control policies for autonomous driving using Real-Time Recurrent Reinforcement Learning (RTRRL), a memory-efficient algorithm that updates policy parameters at every time step without backpropagation through time. We extend RTRRL to support LrcSSM, a recently proposed nonlinear diagonal state-space model, and combine offline behavioral cloning with online RTRRL fine-tuning to adapt policies to distribution shifts at deployment. We validate the approach in the CarRacing simulation and on a 1:10-scale RoboRacer platform equipped with an event camera, where a pretrained policy is fine-tuned online during real-world line-following. To our knowledge, this is the first demonstration of online RL fine-tuning with event-camera observations on standard (non-spiking) hardware in closed-loop control. LrcSSM-based policies improve fastest and most consistently across both settings.
\end{abstract}

\section{Introduction}

% Deployment of pretrained policies often comes with difficulties such as distribution shift due to changes in the environment, sensor drift, etc. Online fine-tuning using RTRRL can mitigate such issues while also adjusting to a changing goal (reward function).

% RTRRL was initially introduced for CT-RNNs and LRUs but can naturally make use of LRCs. We show how gradients can be computed online for fully connected LRCs using RFLO/e-prop, and for diagonalised LRCs using RTRL – with similar complexity.

% We pretrain a policy via behavioral cloning on an offline dataset and then fine-tune it online using RTRRL. We simulate sensor-drift and changes to the reward function – and show how RTRRL does naturally adjust to the changes. This achieves a policy that maintains its performance, while a fixed policy will detoriate.

% Deploying pretrained policies in real-world applications presents substantial challenges that fundamentally limit the practical applicability of learning-based control systems. The core difficulty lies in the distributional mismatch between the training environment and the deployment context, a phenomenon known as distribution shift. When autonomous systems encounter environmental changes in system dynamics, sensor drift, or task objectives, fixed policies rapidly degrade in performance, making them unsuitable for long-term deployment. This is a challenge in robotics and autonomous driving especially, where environmental conditions continuously evolve and mission objectives may change over time.
Learning-based control policies often struggle when deployed outside their training environments. Subtle differences in system dynamics, or sensor characteristics – referred to as distribution shifts – can cause pretrained policies to perform poorly or even fail entirely~\cite{Cobbe2018QuantifyingGI, witty2021measuring, korkmaz2024survey}. This vulnerability of learning-based control policies limits their practical use in autonomous agents that operate in real-world settings – particularly in robotics and autonomous driving – where environmental conditions are constantly evolving, and mission objectives may change over time~\cite{tobin2017domain, voogd2023reinforcement,li2024platform}.

The traditional offline learning paradigm – in which policies are trained once and deployed without further adaptation – proves fundamentally inadequate for handling such non-stationary environments~\cite{levine2020offline}. In such scenarios, online fine-tuning is required: the ability of deployed policies to continuously adapt to changing conditions by updating their parameters in real-time (ideally at each time step), without requiring retraining from scratch or access to large offline datasets. %Online fine-tuning can achieve two critical objectives: (1) mitigating performance degradation caused by distribution shift or sensor drift, and (2) adapting to changes to the reward function that reflect new mission objectives or safety constraints.

\begin{figure}[t]
    \centering
    \includegraphics[width=.45\linewidth]{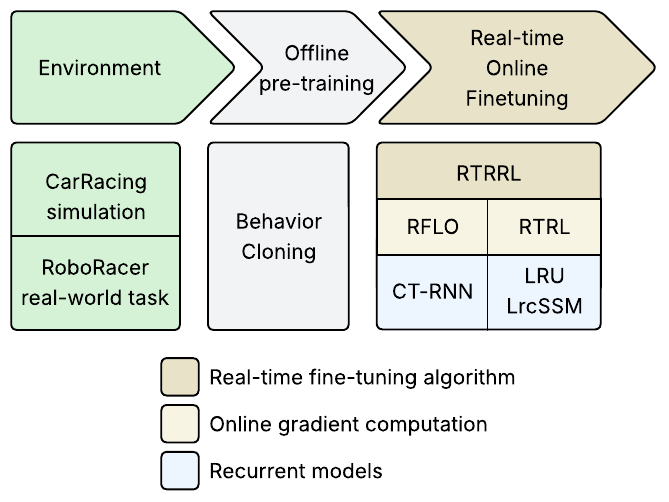}
    \hspace{3em}
    \includegraphics[width=0.3\linewidth]{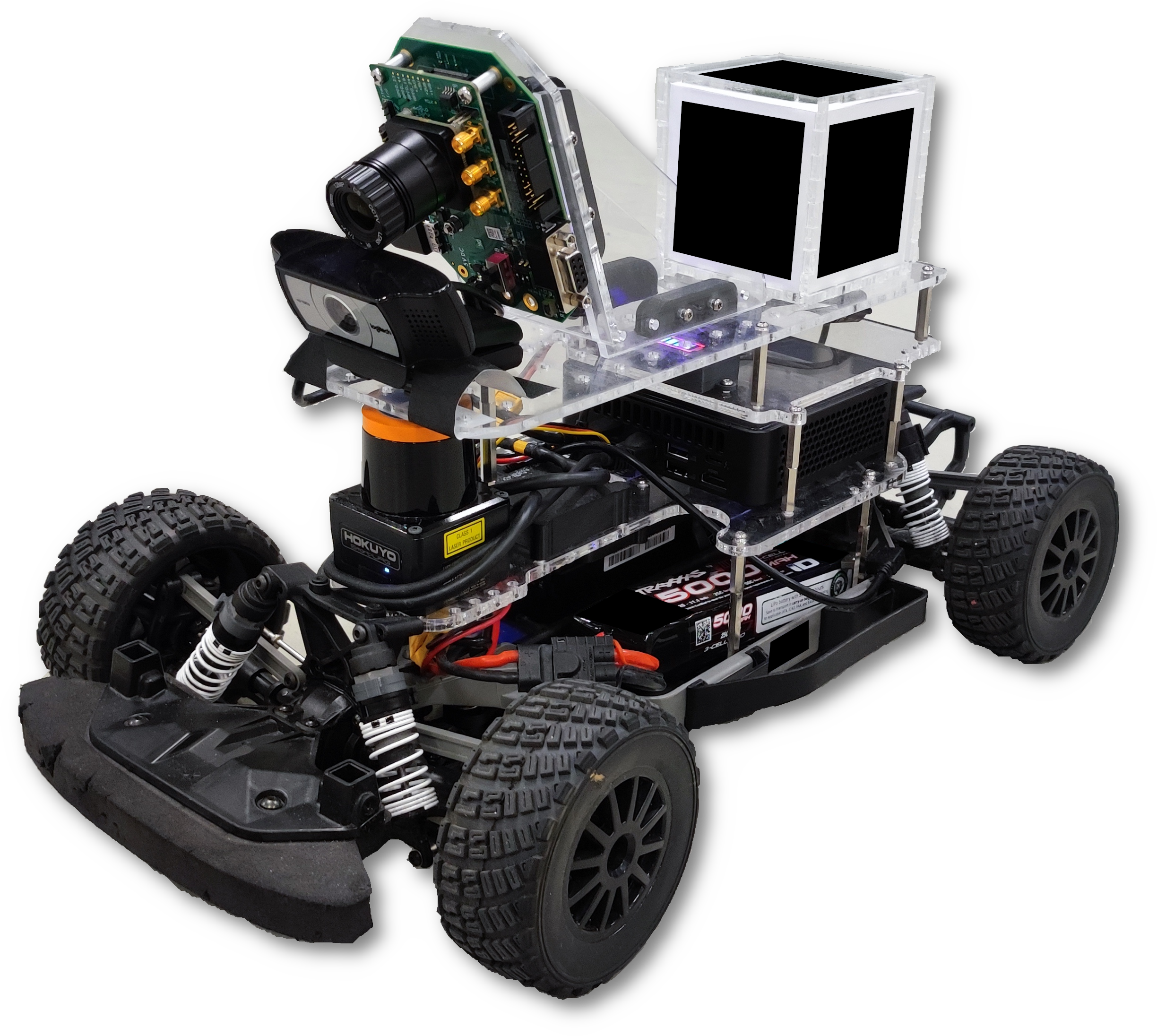}
    \caption{Left: Overview of our proposed method and experiments. After collecting human control data in the environment, a policy is pretrained using behavioral cloning. The policy is then fine-tuned online using RTRRL. The gradients needed for optimization are computed with RTRL or RFLO for diagonalized or fully connected RNN models, respectively. Right: RoboRacer car equipped with a Sony/Prophesee IMX636 sensor for the real-world deployment of the proposed algorithm.}
    
    \label{fig:rtrrl_overview}
\end{figure}

Real-Time Recurrent Reinforcement Learning (RTRRL) \cite{lemmel2025real} offers a solution to this deployment challenge. RTRRL was initially introduced to enable online parameter updates in recurrent neural network policies, supporting both Continuous-Time Recurrent Neural Networks (CT-RNNs) \cite{CTRNN} and Linear Recurrent Units (LRUs) \cite{orvieto2023}. The framework naturally extends to Liquid-Resistance Liquid-Capacitance networks (LrcSSMs) \cite{farsang2024}, a class of bio-inspired continuous-time neural network models that possess desirable stability and expressivity properties for control tasks. Critically, RTRRL computes gradients online using biologically plausible updates. This gradient computation method avoids memory-intensive backward passes, required by backpropagation through time (BPTT).%, making it suitable for deployment on computationally constrained embedded systems.

In this work, we combine offline behavioral cloning with online RTRRL fine-tuning to create adaptive policies that maintain performance under distribution shift. 
%We adopt both simulation- and real-world–based experimental frameworks to systematically evaluate policy adaptation under two classes of distribution shift: (1) sensor drift, simulating realistic degradation of sensor calibration and accuracy, and (2) reward function changes, reflecting scenarios where mission objectives or optimization criteria are modified during deployment. 
Our experimental results demonstrate that RTRRL-based online fine-tuning naturally adjusts policy behavior to maintain task performance in the face of distribution shift, whereas fixed policies deteriorate.
We demonstrate our approach in the \texttt{CarRacing} simulation environment and on a 1:10-scale real-world line-following application. While both of our experimental applications fall within the domain of autonomous driving, the \texttt{CarRacing}-environment application uses conventional RGB-based observations, whereas the real-world lane-keeping application uses a very different event camera. 

% Our work highlights a well-rounded, bio-inspired approach that combines bio-inspired sensors and bio-inspired RNNs, together with a biologically plausible online fine-tuning algorithm. In addition – to the best of our knowledge – this is the first high-frequency closed-loop application of a machine learning algorithm with an event camera on standard (non-spiking) hardware.

\textbf{Contributions:} Our main contributions in this paper are the following:
\begin{itemize}
    % \item \textit{Extending successfully Real-Time Recurrent Reinforcement Learning (RTRRL)} with a novel bio-inspired non-linear state-space model, namely the LrcSSM.
    \item \textit{Integrating LrcSSM into the Real-Time Recurrent Reinforcement Learning (RTRRL) framework} as a drop-in replacement for the LRU backbone, retaining exact RTRL gradient computation while gaining the expressive power of nonlinear state dynamics. To our knowledge, this is the first application of LrcSSM in an online RL setting.
    \item \textit{Proposing for the first time a hybrid learning pipeline} combining offline behavioral cloning with online real-time RTRRL-based fine-tuning, allowing deployed policies to adapt at each time step to non-stationary environments, without retraining from scratch.
    \item \textit{Validating the proposed approach} both in simulation and in a real-world 1:10-scale autonomous driving platform, which highlights the feasibility of biologically-plausible real-time online learning algorithms in embedded robotic systems.
    \item \textit{Presenting the first high-frequency non-spiking machine learning application} of online fine-tuning using event camera observations in a closed-loop control setting. % - to the best of our knowledge.
    \item \textit{Showcasing for the first time the synergy between} bio-inspired sensors, bio-inspired recurrent neural networks, and biologically plausible learning rules - to the best of our knowledge.
    % \item Integrating online fine-tuning into a simulation environment and into real-world scenario, demonstrating the effectiveness of RTRRL.
    % \item First high-frequency non-spiking ML application for event cameras
    % \item ...
\end{itemize}

\section{Related Work}

\paragraph{Interactive Imitation Learning.}
Behavioral cloning can lead to unsatisfactory performance when deploying the novice, due to overfitting on the training set. The DAgger algorithm \cite{ross2010} provides a solution to this problem that relies on interactively extending the initial dataset by allowing the pretrained novice to act at times during task execution, with the expert providing action labels for the resulting observations. This leads to the inclusion of trajectories where the expert is forced to recover from suboptimal states. It is important to note, however, that DAgger approaches still rely on offline supervised learning to update the policy parameters.

\paragraph{Existing Approaches to Online Adaptation.} 
Existing approaches to online adaptation in deployed policies span several paradigms that differ fundamentally from our real-time recurrent approach.
Meta-learning trains policies to rapidly adapt to new tasks or dynamics via few-shot gradient updates or context inference~\cite{finn2017model}. This requires task-specific training data and often suffers from meta-overfitting in continuous control settings~\cite{zhu2025efficient}.
Test-time adaptation techniques enable parameter updates during inference using self-supervised objectives or reward shaping~\cite{xiao2024beyond,liu2026fly}, and recent work in robotics applies test-time RL to vision-language action models for on-the-fly policy refinement in unseen environments~\cite{liu2026fly,xutest}. These methods, however, rely on hand-crafted reward functions or self-supervised objectives rather than rewards provided by the environment. 
%Online label shift adaptation from supervised learning dynamically reweights training examples to match evolving test-time label distributions~\cite{wu2021online}. More general approaches like balanced offline-online replay address full state-action shift ~\cite{lee2021addressing}.
% something like meta-learning (which is obviously not what we do), test-time adaptation~\cite{xiao2024beyond} TODO: add papers from here, adaptive control (more control theory side)~\cite{landau2011adaptive} TODO: stuff from here. Online label shift adaptation~\cite{wu2021online} for classification tasks
%
\paragraph{Online Reinforcement Learning Methods.} 
Conventional approaches for fine-tuning policies typically perform rollouts of full episodes with the current policy, store the collected transitions in a replay buffer, and then update the policy by sampling mini-batches from the buffer at discrete training intervals~\cite{chentowards}. This paradigm introduces an intrinsic latency between interaction and learning, since gradient updates occur only after sufficient experience has been accumulated. 
%Alternatively, the distribution shift problem can be addressed by fine-tuning on offline datasets, maintaining separate replay buffers for offline and online data and carefully balancing the sampling ratio between them during policy updates~\cite{lee2021addressing}.
Compared to these approaches, Real-Time Recurrent Reinforcement Learning (RTRRL)~\cite{lemmel2025real} performs parameter updates at every time-step in a fully online manner, eliminating the need to wait for episodes to complete or to form mini-batches, thereby reducing update latency and enabling strictly causal, fine-grained adaptation in partially observable environments. Compared to the original RTRRL method, our work uses separate RNNs for the policy and critic rather than a shared RNN and adds a CNN encoder, enabling the framework to operate directly on pixel observations. %On top of this, we also integrate LrcSSMs into the framework.
\paragraph{Online Gradient Computation.}
A central challenge for fully online learning in recurrent networks is computing (approximate) gradients in a causal, memory-efficient way. Real-Time Recurrent Learning (RTRL)~\cite{williams1989learning} computes the exact gradient of the loss with respect to all recurrent weights in a strictly forward manner but at the cost of $O(n^{3})$ in the number of units $n$, which makes it impractical for large networks~\cite{williams1989learning,catfolis1993method}. To address this, e-prop~\cite{bellec2020solution} decomposes the gradient into products of neuron-specific eligibility traces and top-down learning signals, providing an online, local learning rule for spiking and rate-based recurrent networks that approaches BPTT performance, while being far more suitable for neuromorphic hardware~\cite{bellec2020solution,rostami2022eprop}. Similarly, Random Feedback Local Online Learning (RFLO)~\cite{10.7554/eLife.43299} derives an approximate gradient rule for RNNs that is local in space and time, and replaces exact backpropagated errors with fixed random feedback weights, yielding a biologically plausible, online update rule of $O(n^{2})$ that performs competitively with BPTT on short horizon tasks~\cite{10.7554/eLife.43299,catfolis1993method, marschall2020}. RTRL is equally efficient with models of diagonal connectivity, as demonstrated in~\cite{zucchet2023}.
RTRRL builds on memory-efficient RTRL with step-wise gradient propagation to update the policy parameters at every time step directly from the incoming reward signal, thereby inheriting the strict online character while tailoring the update rule to temporal-difference RL, rather than a supervised sequence prediction~\cite{lemmel2025real}. 
%
% Random Feedback Local Online Learning (RFLO)~\cite{10.7554/eLife.43299}, 
% e-prop~\cite{bellec2020solution}, 
% Real-Time Recurrent Learning (RTRL)~\cite{williams1989learning}.
%
\paragraph{Event-Based Vision in Machine Learning}
Event-based vision leverages neuromorphic sensors for high-speed, low-latency processing in robotics and control, building on early DVS hardware and algorithms for optic flow and tracking~\cite{gallego2022event, maqueda2018event}. Deep learning advancements, including transformers such as GET~\cite{peng2023get} and comprehensive benchmarks, have been effectively applied to object detection and deblurring in sparse event streams~\cite{zheng2023deep}. Its applications in autonomous vehicles~\cite{9129849} often employ dedicated neuromorphic chips that run spiking neural networks~\cite{9560881, 10342437,paredes2024fully}. In contrast, in this work, we target classical hardware and employ efficient real-time recurrent models, demonstrating that they can also operate on DVS data.
% https://onlinelibrary.wiley.com/doi/full/10.1155/2021/6689337
% https://www.mdpi.com/1424-8220/26/1/81
% https://wires.onlinelibrary.wiley.com/doi/abs/10.1002/widm.1310?casa_token=nVStIzI35isAAAAA%3AKwmXxdAYjtmd0rprgM7PXub-WdsLdyCDmB5XkgTa66mOo0Y_yoFHVbV7GKGYOrvKCk9prMiZ_2gE2_dA
% https://arxiv.org/pdf/2302.08890
% https://ieeexplore.ieee.org/abstract/document/9129849?casa_token=Cb6pNdOAXSgAAAAA:T-o8brGnvEPDZYiQrare22_B_8gU_yleoZkHEm6NiJacQ61dFBTDxNUhKvN4Y3by6nwpHUoJxA
% https://openaccess.thecvf.com/content_cvpr_2018/papers/Maqueda_Event-Based_Vision_Meets_CVPR_2018_paper.pdf
% https://arxiv.org/pdf/2411.03303
% https://ieeexplore.ieee.org/abstract/document/11294255?casa_token=yc7m3A_dL68AAAAA:RRE9yCJQ6X28zGsDd4Mj_d2RGF89WfZE3X8XrT2_FHhwViAZZUTSIL6ObrSkbCWtn-exMwFJBg
% https://ieeexplore.ieee.org/abstract/document/9560881?casa_token=0lDL19m_Pb4AAAAA:ZQLkMZN-U67czltchQdOOPMzXIUkziftGn2AtdyODD-SC5oRuZ5Vkr8tnDvcWjxxrIZJqw7hYw
% https://openaccess.thecvf.com/content/ICCV2023/html/Peng_GET_Group_Event_Transformer_for_Event-Based_Vision_ICCV_2023_paper.html
% https://link.springer.com/chapter/10.1007/978-3-319-70136-3_12
% https://www.frontiersin.org/journals/neurorobotics/articles/10.3389/fnbot.2018.00004/full
% https://ieeexplore.ieee.org/abstract/document/10342437?casa_token=HoVqkjkbVDcAAAAA:slOblhKxUgAgdtXsuh1da4waREnLuPoQ2E-tQC6UPXcb1MzXP5ZiAZ20MaTjPDSoXWaHDZr_yA
% https://ieeexplore.ieee.org/abstract/document/9490245?casa_token=exkOHu__adoAAAAA:FuHTmZXdYyKMq7i3FT-klLGsTQbld8OJHbhd8ggilhX5Z755yDEJVniK4AsQlkoymhRhQ_EknQ
%
\section{Background}
In this section, we briefly describe the methods used in our experiments. We first introduce behavioral cloning for offline training, then present online learning for adaptive improvement using RTRRL, and finally describe the integration of recurrent models into our framework.
\paragraph{Behavioral Cloning}

Behavioral cloning is a simple strategy for learning a controller from expert demonstrations: a dataset of trajectories is collected and used to train a policy offline via supervised learning. The policy can be either \textit{deterministic} or \textit{stochastic}, which determines the loss function used for training. While a deterministic policy predicts actions directly – without any notion of confidence - stochastic policies output probability distributions, such as a Gaussian, which consist of a mean and a standard deviation for each action component.

\paragraph{Online Learning via RTRRL}
Real-Time Recurrent Reinforcement Learning (RTRRL)~\cite{lemmel2025real} is a biologically plausible online learning framework using recurrent neural networks. RTRRL combines online temporal-difference learning with eligibility traces (TD($\lambda$)) with efficient online gradient computation via RTRL or RFLO. The framework performs all learning and weight updates in a single forward pass, without separate backward passes, making it more computationally efficient and biologically realistic than backpropagation through time (BPTT). In the next subsections, we provide brief summaries for the individual building blocks of RTRRL, and we present the algorithm in Appendix~\ref{alg:RTRL}. For further algorithmic details, we refer the reader to~\cite{lemmel2025real}.

\paragraph{RTRRL RNN Architecture}
RTRRL employs a shared RNN backbone with linear output heads for actor $\pi$ and critic $\hat{v}$ functions. At each timestep, the RNN receives observation $o_t$
%, previous actions $a_{t-1}$, and rewards $r_t$
as input, together with the approximate Jacobian $\hat{J}_{t-1}$ computing a latent state representations $h_{t}$ and updated Jacobian $\hat{J}_{t}$: $h_t, \hat{J}_{t}=\mathrm{RNN}_{\theta}(o_t,h_{t-1}, \hat{J}_{t-1})$. We depart from this architecture and employ separate RNNs for each function, yielding distinct latent states $h_{A,t}$ and $h_{C,t}$. The actor network then computes an action distribution $\pi(a|h_{A,t})$ via a linear output layer, and the critic computes a value estimate $\hat{v}(h_{C,t})$. 
%This architecture exploits the meta-RL principle: $\theta_A$,$\theta_C$ weights adapt slowly through training, while hidden states through $\theta_R$ update rapidly at each step. 
%RTRRL uses CT-RNNs or LRUs rather than LSTMs used in~\cite{wang2018}, as their structure supports efficient online gradient computation via RTRL and RFLO. 

\paragraph{Temporal-Difference Learning with Eligibility Traces}
RTRRL uses temporal-difference learning~\cite{sutton2018} combined with eligibility traces (ETs) to enable sample-efficient online updates that handle delayed rewards. After each action, the reward $r_t$, and past and current states $h_{C,t}$ and $h_{C,t+1}$ are used to compute the TD-error $\delta_t$:
\begin{equation}
\delta_t= r_t + \gamma \hat{v}(h_{C,t+1}) - \hat{v}(h_{C,t})
\end{equation}
%
%The value-function $\hat{v}_{\theta_C}(h)$ is learned by regression towards the bootstrapped target. Accordingly, updates are computed by taking the gradient of the value-function and multiplying with the TD-error $\delta_t$: $\theta_{C,t+1} \gets \theta_{C,t} + \eta \delta_t \nabla_{\theta_C} \hat v(h_t)$, where $\eta$ is a small step size. 
%
While standard TD(0) struggles with delayed rewards, RTRRL addresses this through TD($\lambda$), which maintains decaying traces of each parameter's gradient history.
Here, the ET $e_\theta$ decays with factor $\gamma\lambda$ where $\gamma$ is the discount factor. The eligibility trace for the critic $\hat{v}$ is computed as:
\begin{equation}
  e_{\theta_C,t} = \gamma\lambda_C\ e_{\theta_C,t-1} + \nabla_{\theta_C} \hat{v}(h_{C,t})
\end{equation}
%
%RTRRL uses a linear value-function $\hat v_{\theta_C} (h_{t})\,{=}\, w^\top h_{t}$ with parameters $w$, like in the original TD($\lambda$), the gradient of the loss with respect to $w$ is simply $\nabla_{w}\hat v_{\theta_C}\,{=}\,h_{t}$.
%
%We train the Actor and Critic using TD($\lambda$) and take small steps in direction of the log of the action probability $\pi[a]=\mathbf P\mathcal{N}(a|\theta_A^\top h')$ and value estimate $\hat v=\theta_V^\top h'$ respectively. The gradients for each function are accumulated using eligibility traces $e_{A,C}$ with $\lambda$ decay. Additionally, the gradients are passed back to the RNN through random feedback matrices $B_A$ and $B_C$ respectively. The eligibility trace $e_{R}$ for the RNN summarizes the combined gradient. The use of randomly initialized fixed backwards matrices $B_{A,C}$ in RFLO makes RTRRL biologically plausible.
%
In TD($\lambda$), the ET for the actor $\pi$ is computed by taking a small step in the direction of the log of the action probability:
\begin{equation}
\begin{aligned}
     \pi_t &= \pi_{\theta_A,t}(h_t) \qquad a_t \sim \pi_t \\
  e_{\theta_A,t}& = \gamma\lambda_A\ e_{\theta_A,t-1} + \nabla_{\theta_A} \mathrm{log}\ \pi_{t}[a_t]
\end{aligned}
\end{equation}
%
%the gradients are passed back to the RNN through random feedback matrices $B_A$ and $B_C$ respectively. The eligibility trace $e_{R}$ for the RNN summarizes the combined gradient.
Since we use recurrent networks for modeling the policy and value functions, the final eligibility traces additionally incorporate the Jacobian approximations $\hat{J}_t$:
\begin{equation}
\begin{aligned}
  e_{\theta_C,t} &= \gamma\lambda_C\ e_{\theta_C,t-1} + \hat{J}_{C,t} \nabla_{h_{C,t}} \hat v(h_{C,t})\\
   e_{\theta_A,t} &= \gamma\lambda_A\ e_{\theta_A,t-1} + \hat{J}_{A,t} \nabla_{h_{A,t}} \mathrm{log}\ \pi_{t}[a_t]
   \end{aligned}
\end{equation}
Once the eligibility traces and the TD-error $\delta_t$ are computed, the parameters $\theta_C$ and $\theta_A$ are updated using a small step $\theta_{t+1} \leftarrow {\theta_t} + \eta\ \delta_t\ e_{\theta,t}$ with corresponding $\eta$ step size and $e_\theta$ eligibility trace.

\paragraph{Online Gradient Computation}
RTRRL's key innovation is computing RNN gradients online without BPTT, using Real-Time Recurrent Learning (RTRL) or Random Feedback Local Online (RFLO). Both methods maintain a Jacobian trace updated during forward computation:
\begin{equation} 
\label{eq:RTRL}
\hat J_{t+1} =  \left( \mathbb I + \nabla_{h_{t}} f(x_{t},h_{t})\right) \hat J_{t} + \bar J_t
\end{equation}
allowing gradient updates at each step. However, RTRL has $O(n^3)$ memory complexity in neuron count $n$, making it impractical for large networks. 
RFLO approximates RTRL for CT-RNNs with $O(n^2)$ complexity by enforcing biological plausibility: it avoids weight transport by using fixed random feedback matrices $B$ (feedback alignment), and enforces locality by dropping non-local gradient terms:
\begin{equation} 
\label{eq:RFLO}
\hat J_{t+1}^W \approx (1 -\tau^{-1})\ \hat J _t^W  + \tau^{-1} \varphi'(W \xi_t) \xi_t^\top
\end{equation}
The update is $\Delta W(t) =\hat J_t^W B \varepsilon_{t}$ using a fixed random matrix $B$. Prior work shows that this simplified formulation remains effective for learning \cite{murray2019, marschall2020}.

\subsection{Recurrent Neural Network Models}
Here, we provide a brief overview of the models that we integrate into the RTRRL framework. 

\paragraph{CT-RNN.}
Introduced in \citep{CTRNN}, continuous-time recurrent neural networks (CT-RNNs) can be interpreted as electrical-equivalent circuits that model electrical synapses~\cite{gerstner2014, farsang2024learning}. Their dynamics are described by the following ordinary differential equation:
%In its condensed form, a CT-RNN with $N$ hidden units, $I$ inputs, and $O$ outputs has the following latent-state dynamics:
%
\begin{equation}
\label{eq:ctrnn}
% h_{t+1} = h_{t}+\frac{1}{\tau}\left(-h_{t}+\varphi(W \xi_t)\right) \quad \xi_t=\begin{bmatrix}  x_{t} \\h_{t}\\ 1\end{bmatrix} \in \R^{Z}
\dot h(t) = \frac{1}{\tau}(-h(t)+\varphi(W \xi(t)))
\end{equation}
where $h \in \mathbb{R}^N$ denotes the hidden state, $\varphi(\cdot)$ is a nonlinear activation function, and $\tau \in \mathbb{R}^N$ is a vector of neuron-specific time constants. The augmented input $\xi \in \mathbb{R}^{I+N+1}$ concatenates the external input $x(t)\in \mathbb{R}^{I}$, the current hidden state at time step $t$, and a constant bias term. The weight matrix $W \in \mathbb{R}^{N \times (I+N+1)}$ jointly parameterizes input, recurrent, and bias connections. The network output is obtained via a linear readout, $y(t)=W_{out}h(t)$.
% where $x_{t}$ is the input at time $t$, $\varphi$ is a non-linear activation function, $W$ a combined weight matrix $\in \R^{N\times X}$, $\tau$ the time-constant per neuron $\,{\in}\,\R^{N}$, and $\xi$ a vector $\in \R^{Z}$ with $Z\,{=}\,I\,{+}\,N\,{+}\,1$, the $1$ concatenated to $\xi_t$ accounting for the bias. The output $\hat y_{t} \in \R^O$ is given by a linear mapping $\hat y_{t}=W_{out}h_t$. The latent state follows the ODE defined by $\dot h_t = \tau^{-1}(-h_{t}+\varphi(W \xi_t))$, an expression that tightly resembles conductance-based models of the membrane potentials in biological neurons~\cite{gerstner2014}. 

\paragraph{LRU.}

% The name Linear Recurrent Unit was introduced in the seminal work of \citet{orvieto2023}. LRUs gained a lot of attention recently as they were shown to perform well in challenging tasks. The linear recurrence means that updates can be computed very efficiently. 
% As a special case of State-Space Models \citep{gu2021}, the latent state of this simple RNN model is described by a linear system:
%
Linear Recurrent Units (LRUs)~\citep{orvieto2023} – describe their latent dynamics through a linear system, enabling highly efficient state updates. They define the following state-space model (SSM) equations:
\begin{equation}
\label{eq:lru}
% \dot h_t=A h_t+B x_t \qquad \quad y_t=\Re\left[C h_t\right]+D  x_t
\dot h(t)=A h(t) + B x(t)
\end{equation}
%
% where $A$ is a diagonal matrix $\in \C^{N\times N}$ and $B, C, D$ are matrices $\in \C^{N\times I}, \C^{O\times N}$ and $\R^{O\times I}$ respectively. Note that the hidden state $h_t$ is a complex-valued vector $\in \C^N$ here. For computing the output $y_t$, the real part of the hidden state is added to the input $x_t$ at time $t$.
where $A \in \mathbb{C}^{N \times N}$ is a diagonal matrix, and $B \in \mathbb{C}^{N \times I}$, $C \in \mathbb{C}^{O \times N}$, and $D \in \mathbb{R}^{O \times I}$ are learnable parameter matrices. The hidden state $h_t \in \mathbb{C}^N$ is complex-valued. The output $y(t)=\Re\left[C h(t)\right]+D  x(t)$ is obtained by combining the real part of the latent state with the input $x(t)$.

\paragraph{LRC.}
Liquid-Resistance Liquid-Capacitance models (LRCs)~\cite{farsang2024} and their diagonal variant, LrcSSMs~\cite{farsang2025parallelizationnonlinearstatespacemodels}, have an intermediate position between CT-RNNs and LRUs. From a biological perspective, LRCs are models of chemical synapses, whereas CT-RNNs can be interpreted as models of electrical synapses. A key distinction between the two lies in their treatment of temporal dynamics: CT-RNNs employ fixed time constants, whereas LRCs introduce input- and state-dependent time constants, allowing each neuron to adapt its temporal behavior.

From a state-space modeling viewpoint, both LRUs and LrcSSMs fall within the SSM framework. While LRUs are characterized by purely linear recurrence, LrcSSMs extend this formulation by introducing nonlinear state dynamics. Despite this increased expressiveness, LrcSSMs retain computational efficiency due to their diagonal structure, which enables decoupled and parallelizable state updates.
The dynamics of LRCs can be expressed as
\begin{equation}
\label{eq:lrc}
\dot h(t) = A_{h,x}(t)\, h(t) + b_{h,x}(t)
\end{equation}
where $A_{h,x}(t) \in \mathbb{R}^{N \times N}$ is a diagonal matrix and $b_{h,x}(t) \in \mathbb{R}^{N}$ is a vector-valued function, capturing biologically interpretable, state- and input-dependent dynamics.

\section{Experimental Setup}
\subsection{Integrating LrcSSM into online fine-tuning framework}
% now: bio-inspired (CT-RNN ~ electrical synapse model, LRC ~ chemical synapse), diagonal efficient version (LRU ~ linear, LRC ~ non-linear)
%Priority to RTRRL -> diagonal LRC version

%The biological plausibility of RTRRL stems form the fact that BPTT is avoided when computing gradients for the recurrent neural network. 
As mentioned above, gradients for recurrent models with diagonal connectivity can be computed exactly – and without a loss in performance – using RTRL (Eq. \ref{eq:RTRL}). 
The diagonalized version of LRC (LrcSSMs) can be used as a drop-in replacement of LRU in the original RTRRL algorithm. The resulting algorithm retains the benefits of cheap computation of exact gradients, while also regaining the biological interpretation that the LRU model lacks.

\subsection{Model Architecture}

\begin{wrapfigure}{R}{0.56\linewidth}
    \centering
    \includegraphics[width=\linewidth]{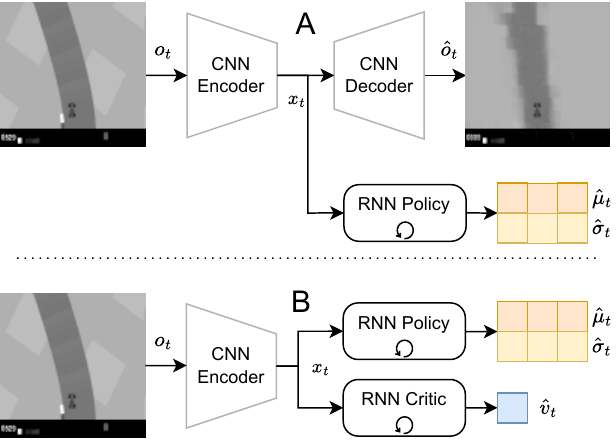}
    \caption{Model architecture. A: For pretraining, a CNN autoencoder is used to encode the image observation $o_t$; the encoded vector $x_t$ serves as input to the RNN policy predicting action distribution $\mathcal N(\hat \mu_t, \hat \sigma_t)$. B: The RNN policy is fine-tuned; to this end, an RNN critic is used that predicts value estimate $\hat v_t$ from encoded vector $x_t$.}
    \label{fig:model}
    \vspace{-1em}
\end{wrapfigure}

Figure \ref{fig:model} shows the model structure used for our experiments. The core components are the convolutional encoder and the recurrent policy, which are pretrained first with supervised learning and later fine-tuned with reinforcement learning. The convolutional decoder and the recurrent critic are used only during pretraining and fine-tuning, respectively.

During pretraining, the convolutional encoder is part of an autoencoder that is trained to minimize the image reconstruction loss $\mathcal L_{rec} = ||o_t - \hat o_t||$. The actor is modeled as a single-layer recurrent neural network and receives the encoded observation $x_t$ as input. Its output is transformed by a fully connected layer that maps to a number of outputs equal to double the action dimension. The outputs correspond to the mean and standard deviation of a Gaussian distribution that is transformed by a hyperbolic tangent bijector and scaled to the maximum action magnitude afterward. This setup ensures that the policy can only predict admissible actions. Training of the policy is facilitated through maximizing the log-probability of the action in the dataset under the predicted distribution $\mathcal L_{act} = -\log P(a_t|x_t)$. The combined loss for pretraining is the sum of the action prediction loss and the weighted reconstruction loss: $\mathcal L = \mathcal L_{act} + \eta_{rec} \mathcal  L_{rec}$.

For fine-tuning via RTRRL, a recurrent value function is randomly initialized and used alongside the CNN encoder and the RNN policy. The RNN Critic receives the same encoded vector $x_t$ as input, and its output is mapped by an all-to-one fully connected layer. The Critic's output corresponds to a deterministic prediction of the expected cumulative state-value with discounting.
Contrary to \citep{lemmel2025real}, we use separate RNNs for the policy and the critic; that is, we do not share the RNN encoder.

\subsection{Parameter Change Penalty}

When fine-tuning a policy on a single environment with an effective batch size of $1$, overfitting is likely to appear. To combat this, we add an additional penalty to the policy loss that penalizes the $L_2$ norm of the difference between pretrained and current policy parameters, similar to \citep{lemmel2025a}.
\begin{equation}
    L_\theta = \beta\ ||\theta_{pre} - \theta_t||_2
\end{equation}
The gradient of the penalty with respect to the policy parameters is computed using RTRL/RFLO and added to the gradients resulting from TD($\lambda$).

\subsection{Simulation Environment}

For our simulation experiments, we introduce a modified version of the \texttt{CarRacing} environment from OpenAI Gym by augmenting it with an additional penalty term that penalizes deviations of the vehicle’s wheels from the track centerline.
Specifically, each segment of the track is defined by a keypoint representing its center of mass. We compute the Euclidean distance of the car's center to all keypoints of the track and take the minimum of these distances. This deviation measure is scaled and subtracted as a penalty term from the original environment reward.
The modified reward encourages agents to maintain a trajectory that keeps the vehicle’s wheels close to the track centerline, effectively shaping the policy toward safer, more comfortable driving behavior. %As a result, the findings are not directly comparable to the standard CarRacing benchmark.

We first collected human demonstration data by letting an average human player complete three laps in the environment. Each trajectory consists of a sequence of 1000 image observations, each paired with the corresponding control action. This dataset is the source of supervised pretraining for the agent.
For each model type, we initialized parameters with five different random seeds and pretrained each on the recorded human trajectories. These models are subsequently used as initializations for the fine-tuning phase, during which the agents continue training in the modified environment.

\subsection{Real-world deployment}

In our real-world deployment, we used a RoboRacer car – an autonomous platform based on a 1:10-scale racecar with an Asus-NUC-based compute platform – equipped with a Sony/Prophesee IMX636 Dynamic Vision Sensor (DVS). The compute platform is connected to a motor and steering controller (see Fig. \ref{fig:rtrrl_overview} right). % and can use additional sensors for control.
Unlike an RGB optical sensor, the DVS captures changes in pixel intensity and generates a stream of intensity-change events, triggered when the intensity exceeds a predefined threshold. 
% Because processing the raw event stream is resource-intensive, most applications, including ours, use a presentation of the event data. 
% Typically, those representations use events within a certain time frame as a basis. 
%The representation closest to the raw event stream is the \textit{event tensor}, which contains all event data within the time frame, usually aggregated along the time dimension, in a single three-dimensional tensor. While this representation allows for the processing of all events, it mitigates a major selling point of event cameras, which is the sparse nature of event data. To reduce the size of the representation, the \textit{time surface} flattens the time axis into a single plane, retaining the movement gradient of the raw events, while providing a two-dimensional representation of the data. 
We use aggregated event data to generate frame-based representations for conventional (non-spiking) neural networks. The representation we use, \textit{event frames}, retains only the polarity of each event per pixel and discards all sub-frame timing information. Figure \ref{fig:representations} in the appendix shows this representation with the corresponding RGB frame.
Typically, filtering is applied to each representation to remove noise from the event stream, and some representations also flatten event polarities. \citep{gallego2022event} describe different representations in detail and typical algorithms for event data.

Our \texttt{LineTracking} experiment uses the dataset collected by~\cite{resch2025mmdvs} from real-world driving sessions conducted by non-expert drivers. The dataset consists of 500772 event frames, generated at 100 Hz, each containing event data representations of size 128 × 64 × 2, paired with the corresponding control actions $\in \mathbb R^3$. 
Similar to the \texttt{CarRacing} setup, this dataset is used to perform supervised pretraining for each model across multiple random seeds before fine-tuning within the simulation environment.

For fine-tuning, we use a reward function based on the agent's distance to the reference line, the heading difference, and the rate of change of the steering angle. To allow for slight deviations, each reward modality uses a slack value to indicate whether the task is achieved, and the reward is scaled by a weight for each modality. %The reward is the sum of the weighted partial rewards.
%
%\begin{align*}
%    r_{m} &= (slack_{m} - |ref_{m} - actual_{m}|) * Q_{m} \\
%    r &= \sum_{m = 0}^M r_m
%\end{align*}
%
To determine the car's pose, we first map the experimental area with \verb|slam_toolbox|\footnote{\url{https://github.com/SteveMacenski/slam_toolbox}} and record key points along the marked line. Using the mapped environment, we estimate the agent's pose via a Monte Carlo-based approach inspired by \cite{walsh17}. %With the pose estimate and the key points, we derive the distance to the line and the heading deviation for the reward function.

We evaluate the robustness of our approach to distribution shifts by shifting the output steering angle by 0.1 rad, which corresponds to $\approx 19 \%$ of the action range, and introducing visual disturbances caused by ambient (sun) light entering our lab.
The steering shift is applied after 100 seconds of inference, typically during the third lap, depending on the model's performance.

\section{Results}

The aim of this paper is to show that models pretrained on a limited data set can be fine-tuned thereafter using RTRRL. Our results show that this approach holds true for both of our experiments. In this section, we first discuss pretraining and fine-tuning results for the \texttt{CarRacing} simulation, followed by our results for the real-world \texttt{LineTracking} application.

\subsection{CarRacing}
\paragraph{Pretraining results}

\begin{figure}
  \begin{minipage}[t]{.56\linewidth}\vspace{0pt}%
  \centering
    \vspace*{-2ex}
    \includegraphics[width=.49\linewidth]{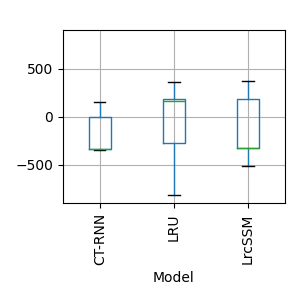}
    \includegraphics[width=.49\linewidth]{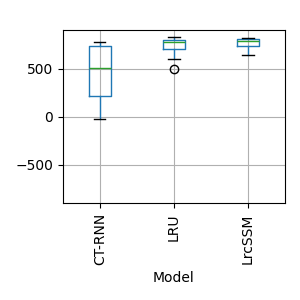}
    \vspace*{-4ex}
    \captionof{figure}{Boxplots of evaluation reward on three different tracks, for five different pretrained models, aggregated per type. Left plot shows rewards before fine-tuning. Right plot shows rewards after fine-tuning.}% \caption{Figure caption}
    \label{fig:carracing_boxplots}
  \end{minipage}\hfill
  \begin{minipage}[t]{.42\linewidth}\vspace{0pt}%
    \centering
    \includegraphics[width=\linewidth]{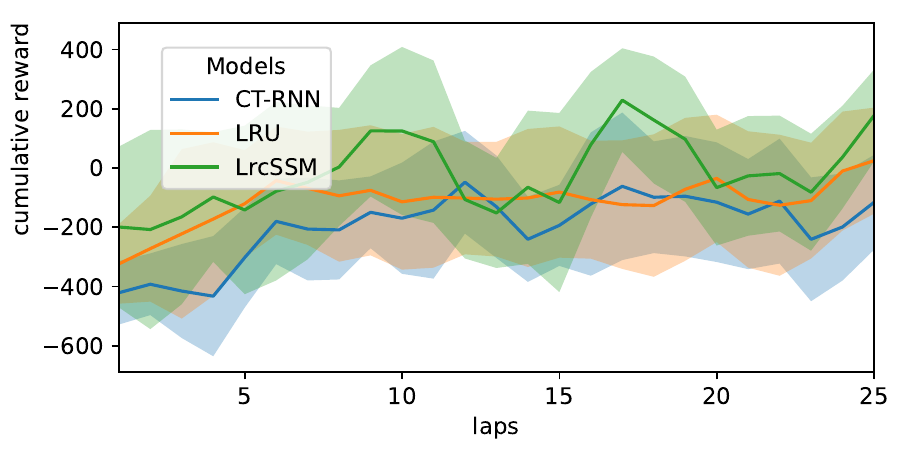}
    \captionof{table}{Lineplot of median evaluation reward on three different tracks of the \texttt{CarRacing} environment for all pretrained models, aggregated per type. Shaded regions depict the standard deviation.}
    \label{fig:carracing_finetuning}
  \end{minipage}
\end{figure}

Figure \ref{fig:carracing_pretraining} in the Appendix shows the validation loss curves for pretraining the \texttt{CarRacing} policies, aggregated by model type. Figure \ref{fig:carracing_boxplots} (left) shows the aggregated evaluation reward for the pretrained models when used in the environment without any tuning. %At this point, all pretrained models performed in a similar fashion.

\paragraph{Fine-Tuning}

We fine-tuned the pretrained models using RTRRL by updating the policy and value functions after each environment step. 
Since RTRRL – the algorithm we use for fine-tuning – is sensitive to hyperparameter choices, we conducted several hyperparameter sweeps to identify a good configuration. We found the actor learning rate to be the most sensitive hyperparameter, with critic learning rate and entropy regularization having minimal impact. Table \ref{tab:hyperparams} in the appendix lists the final hyperparameter configuration for RTRRL that was used for the remainder of the experiments.

Figure \ref{fig:carracing_finetuning} shows the median cumulative lap reward over 10 laps of fine-tuning for 5 random seeds per model type. A clear upward trend is visible across all model types, with LrcSSMs improving the most.
Figure \ref{fig:carracing_boxplots} right shows the aggregated evaluation reward for the pretrained models after 10 laps of fine-tuning. During evaluation, the mode of the predicted action policy is executed rather than sampled from it during fine-tuning. As shown, all models have improved significantly over the initial reward after fine-tuning. Moreover, the LrcSSM results improved the most and showed very consistent results across all models and seeds. 

\begin{figure}
  \begin{minipage}[t]{.56\linewidth}\vspace{0pt}%
    \centering
    \vspace*{2.5ex}
    \includegraphics[width=.49\linewidth]{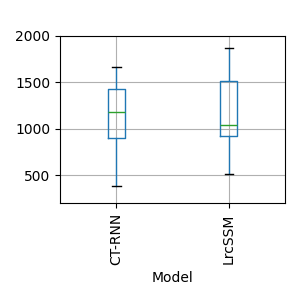}
    \includegraphics[width=.49\linewidth]{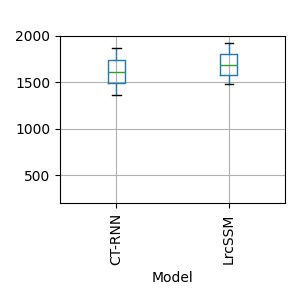}
    \captionof{figure}{Boxplots of cumulative rewards for the \texttt{Line}-\texttt{Tracking} experiment of five different pretrained models, aggregated per type. Left plot shows the rewards before fine-tuning. Right plot shows them after fine-tuning. LRUs were excluded due to their very poor performance after pretraining, which made deployment tedious.}% \caption{Figure caption}
    \label{fig:linetracking_boxplots}
  \end{minipage}\hfill
  \begin{minipage}[t]{.42\linewidth}\vspace{0pt}%
    \centering
    \vspace*{-3ex}
    \includegraphics[width=\linewidth]{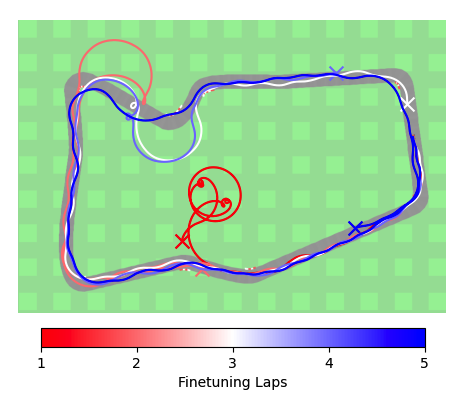}
    \captionof{table}{Shown are the trajectories of five laps of fine-tuning of an originally suboptimal policy. Initially, the car goes off the road (red), but it improves with each lap, eventually completing the track (blue).}
    \label{fig:carracing_laps}
  \end{minipage}
\end{figure}

To illustrate the effect of fine-tuning in the \texttt{CarRacing} environment, we picked one LrcSSM policy for plotting lap trajectories shown in figure \ref{fig:carracing_laps}. While the policy was not able to successfully complete the track at first, it quickly adjusted to it within five laps.

\subsection{Real-world deployment}

For the \texttt{LineTracking} task, the agent was placed on a pre-determined starting point on a line marked on the floor with clearly distinguishable tape. The goal of the task was following the line as closely as possible while avoiding rapid steering inputs.
For evaluation, we chose an arbitrary point on the line as the start/finish line for this task to segment the agent's trajectories into laps. %In this instance, we picked the middle of a short straight, as most policies successfully completed this section of the circuit.

In Figure \ref{fig:carracing_pretraining} of the appendix we show the validation-loss curves for pretraining on the \texttt{LineTracking} dataset, and in Figure \ref{fig:linetracking_boxplots} left we show the boxplot of the cumulative rewards of the pretrained models over three laps and without any fine-tuning.

% \paragraph{Pretraining results}

% Figure \ref{fig:linetracking_pretraining} in the appendix, shows the validation loss curves that were recorded during pretraining of the \texttt{LineTracking} dataset. While LrcSSMs and CT-RNNs achieved comparable results after 2000 epochs, LRUs appeared to be significantly worse. This shortcoming of LRUs became even more evident when deploying the pretrained models to the real-world car, as none of theses models were able to complete a lap without a large number of human interventions. A boxplot of the number of human interventions is shown in Figure \ref{fig:interventions} in the Appendix. We suspect that the linear state-space model semantics of the LRU model is not expressive enough to satisfactorily capture the steering behavior when used as a single-layer RNN. Due to the very poor performance of LRUs, we omitted them from the fine-tuning analysis altogether.

\paragraph{LRU pretraining failure as a finding} While LRUs achieved reasonable validation loss on CarRacing, they failed to learn an effective LineTracking policy: pretrained LRU models could not complete a single lap without numerous human interventions (see Figure \ref{fig:interventions} in the appendix). We hypothesize that this stems from the LRU's purely linear recurrence, which appears insufficient to capture the discontinuous, threshold-like steering behavior required by the event-frame input transitioning abruptly between line-visible and line-absent states. In contrast, LrcSSMs extend the linear state-space formulation with input- and state-dependent nonlinear dynamics while retaining the diagonal structure that enables exact RTRL gradients, allowing them to succeed on the same task with the same architecture and computational budget. This suggests that the nonlinear state dynamics – not the diagonal structure per se – are essential for event-based control. We therefore excluded LRUs from the fine-tuning experiment and report their pretraining failure as a finding.

\paragraph{Fine-Tuning}

\begin{figure}
  \begin{minipage}[t]{.475\linewidth}\vspace{0pt}%
    \centering
    \vspace*{-1ex}
    \includegraphics[width=.9\linewidth]{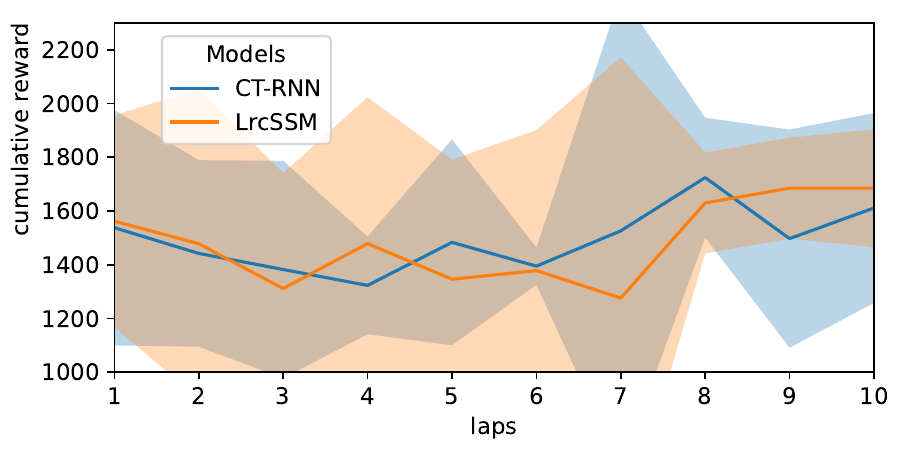}
    \vspace*{-1ex}
    \captionof{figure}{Median cumulative rewards per lap for the \texttt{LineTracking} experiment. Shaded regions are showing the standard deviation. As can be oobserved, LrcSSM exhibit a superior behavior.}% \caption{Figure caption}
    \label{fig:diff_models}
  \end{minipage}\hfill
  \begin{minipage}[t]{.475\linewidth}\vspace{0pt}%
    \centering
    \begin{tabularx}{\textwidth}{X|l|ll}
        \toprule
         & & \multicolumn{2}{c}{\textbf{After Shift}} \\
         \textbf{Shift} & \textbf{Initial} & \textbf{First} & \textbf{Best} \\
         \midrule
         Steering & 1901 & -4854 & -1708 \\
         Steering & 2418 & -2376 & -1064 \\
         Steering \& Sun & 1779 & -1307 & -938 
         % \bottomrule
    \end{tabularx}
    \vspace*{0.2ex}
    \captionof{table}{Cumulative rewards of the CT-RNN model variant to different distribution shifts. Initial cumulative reward is taken from the last un-shifted lap, and the First After Shift, from the first full lap with the shift.}
    \label{tab:distr_shift_real_world}
  \end{minipage}
\end{figure}

For the real-world fine-tuning experiment, we selected the three best pretrained models for LrcSSMs and CT-RNNs to deploy on the autonomous vehicle. A human operator monitored the car during the experiment and intervened as soon as a significant deviation from the track was observed, maneuvering the car back to a point with a clear camera view of the line. 

Figure \ref{fig:linetracking_boxplots} right shows the cumulative rewards achieved by the CT-RNN and LrcSSM models after fine-tuning, and Figure \ref{fig:diff_models} shows the cumulative rewards over 10 laps aggregated per model.
%
%Very pleasantly surprising, we witnessed an immediate improvement in the first lap, hinting at an instantaneous effect of the fine-tuning. From there on, the performance of the models seemed to stagnate at first, followed by a significant reduction in standard deviation over the course of the fine-tuning.
% Addressing 4)
Most models showed an initial decrease in reward, which we attribute to the critic being randomly initialized, but then improved their performance to levels above those of the initial policy.
%
% \begin{figure}
%     \centering
%     \includegraphics[width=\linewidth]{figures/experiments/ctrnn/cumulative_rewards.limit.ovlf.pdf}
%     \caption{Cumulative rewards per lap for CT-RNNs.}
%     \label{fig:fine_tuning_rewards_ctrnn}
% \end{figure}
% \begin{figure}
%     \centering
%     \includegraphics[width=\linewidth]{figures/experiments/lrc/cumulative_rewards.limit.ovlf.pdf}
%     \caption{Cumulative rewards per lap for LRCs.}
%     \label{fig:fine_tuning_rewards_lrc}
% \end{figure}
%
Showcasing the effectiveness of RTRRL on the real-world task, Figure \ref{fig:finetune_laps_lrc} in the appendix shows the trajectories of an LrcSSM policy. While it initially fails to follow the line (lower-left of the figure), it eventually does so in successive laps. Further laps additionally show reduced oscillations around the target line. %Similar results were observed for CT-RNNs.
\subsection{Distribution shifts}
We conducted additional distribution shift experiments in both simulation and the real world: after a certain number of steps $N_s$, shift the steering angle by a fixed amount $\delta_{s}$. We chose $\delta_{s}=0.2$ rad and $0.1$ rad for \texttt{CarRacing} and \texttt{LineTracking} respectively. For both environments, we chose $N_{s}=10000$ – corresponding to 10 laps in the \texttt{CarRacing} simulation and to roughly 3 laps in the real-world line-tracking experiment.
\begin{table}[h]
\centering
\caption{Results for distribution shift experiments. The first two columns show the initial reward directly after pre-training and with applied shift. The third column shows the reward after fine-tuning.}
\vspace*{2mm}
\begin{tabular}{lccc}
\toprule
\textbf{Model name} & \textbf{Initial eval reward} & \textbf{Initial eval shift} & \textbf{After fine-tuning} \\
\midrule
LrcSSM & $-362.68 \pm 243.39$ & $-390.99 \pm 164.79$ & $-87.98 \pm 440.85$ \\
LRU & $180.92 \pm 399.34$ & $-290.46 \pm 433.17$ & $-116.04 \pm 9.81$ \\
CT-RNN & $-229.50 \pm 315.11$ & $-422.73 \pm 390.61$ & $143.16 \pm 502.11$ \\
\bottomrule
\end{tabular}
\label{tab:model_performance}
\end{table}
Table \ref{tab:model_performance} shows the results of the distribution shift experiments in simulation. The fine-tuning was effective in adjusting to the shifted steering angle in simulation. Table \ref{tab:distr_shift_real_world} shows the cumulative rewards of CT-RNNs per lap, with applied distribution shifts in the real world.
In all experiments, the reward achieved by the model first degraded, and then began to improve again, reaching levels comparable to the initial rewards.

% \begin{figure}
%     \centering
%     \includegraphics[width=\linewidth]{figures/experiments/ctrnn/cumulative_rewards.limit.ovlf.intr.pdf}
    
%     \caption{Manual interventions per lap for CT-RNNs.}
%     \label{fig:fine_tuning_interventions_ctrnn}
% \end{figure}

% \begin{figure}
%     \centering
%     \includegraphics[width=\linewidth]{figures/experiments/lrc/cumulative_rewards.limit.ovlf.intr.pdf}
    
%     \caption{Manual interventions per lap for LRCs.}
%     \label{fig:fine_tuning_interventions_lrc}
% \end{figure}

\section{Discussion}

In this paper, we asked whether online fine-tuning of a pretrained policy can be achieved using RTRRL – a biologically plausible forward-only reinforcement learning algorithm. To this end, we created or used a static dataset of human control inputs and employed behavioral cloning to pretrain a combined convolutional-recurrent policy that predicts actions from pixel inputs. We then fine-tuned the policy online using RTRRL – updating parameters after each step – to improve the policy on the fly.
Our experimental results show substantial improvements in the simulated \texttt{CarRacing} environment that carry over to our real-world \texttt{LineTracking} example. 

\section*{Acknowledgement}
J.L. was partially supported by the Doctoral College Resilient Embedded Systems (DC-RES) of TU Wien. F.R. and R.G. have received funding from the European Union's Horizon Europe research and innovation program with Grant Agreement No. 10039070. M.F. has received funding from the European Union’s Horizon 2020 research and innovation programme under the Marie Skłodowska-Curie grant agreement No 101034277. Computational results have been achieved in part using the Vienna Scientific Cluster (VSC).

\bibliography{example_paper}

\begin{thebibliography}{10}

\bibitem{Cobbe2018QuantifyingGI}
K.~Cobbe, O.~Klimov, C.~Hesse, T.~Kim, and J.~Schulman, ``Quantifying generalization in reinforcement learning,'' in {\em International Conference on Machine Learning}, 2018.

\bibitem{witty2021measuring}
S.~Witty, J.~K. Lee, E.~Tosch, A.~Atrey, K.~Clary, M.~L. Littman, and D.~Jensen, ``Measuring and characterizing generalization in deep reinforcement learning,'' {\em Applied AI Letters}, vol.~2, no.~4, p.~e45, 2021.

\bibitem{korkmaz2024survey}
E.~Korkmaz, ``A survey analyzing generalization in deep reinforcement learning,'' {\em arXiv preprint arXiv:2401.02349}, 2024.

\bibitem{tobin2017domain}
J.~Tobin, R.~Fong, A.~Ray, J.~Schneider, W.~Zaremba, and P.~Abbeel, ``Domain randomization for transferring deep neural networks from simulation to the real world,'' in {\em 2017 IEEE/RSJ international conference on intelligent robots and systems (IROS)}, pp.~23--30, IEEE, 2017.

\bibitem{voogd2023reinforcement}
K.~L. Voogd, J.~P. Allamaa, J.~Alonso-Mora, and T.~D. Son, ``Reinforcement learning from simulation to real world autonomous driving using digital twin,'' {\em IFAC-PapersOnLine}, vol.~56, no.~2, pp.~1510--1515, 2023.

\bibitem{li2024platform}
D.~Li and O.~Okhrin, ``A platform-agnostic deep reinforcement learning framework for effective sim2real transfer towards autonomous driving,'' {\em Communications Engineering}, vol.~3, no.~1, p.~147, 2024.

\bibitem{levine2020offline}
S.~Levine, A.~Kumar, G.~Tucker, and J.~Fu, ``Offline reinforcement learning: Tutorial, review, and perspectives on open problems,'' {\em arXiv preprint arXiv:2005.01643}, 2020.

\bibitem{lemmel2025real}
J.~Lemmel and R.~Grosu, ``Real-time recurrent reinforcement learning,'' in {\em Proceedings of the AAAI Conference on Artificial Intelligence}, vol.~39, pp.~18189--18197, 2025.

\bibitem{CTRNN}
K.-i. Funahashi and Y.~Nakamura, ``Approximation of dynamical systems by continuous time recurrent neural networks,'' {\em Neural Networks}, vol.~6, pp.~801--806, Jan. 1993.

\bibitem{orvieto2023}
A.~Orvieto, S.~L. Smith, A.~Gu, A.~Fernando, C.~Gulcehre, R.~Pascanu, and S.~De, ``Resurrecting recurrent neural networks for long sequences,'' in {\em Proceedings of the 40th {{International Conference}} on {{Machine Learning}}}, vol.~202 of {\em {{ICML}}'23}, pp.~26670--26698, JMLR.org, 2023.

\bibitem{farsang2024}
M.~Farsang, S.~A. Neubauer, and R.~Grosu, ``Liquid {{Resistance Liquid Capacitance Networks}},'' in {\em The {{First Workshop}} on {{NeuroAI}} @ {{NeurIPS2024}}}, arXiv, Nov. 2024.

\bibitem{ross2010}
S.~Ross and D.~Bagnell, ``Efficient {{Reductions}} for {{Imitation Learning}},'' in {\em Proceedings of the {{Thirteenth International Conference}} on {{Artificial Intelligence}} and {{Statistics}}}, pp.~661--668, {JMLR Workshop and Conference Proceedings}, Mar. 2010.

\bibitem{finn2017model}
C.~Finn, P.~Abbeel, and S.~Levine, ``Model-agnostic meta-learning for fast adaptation of deep networks,'' in {\em International conference on machine learning}, pp.~1126--1135, PMLR, 2017.

\bibitem{zhu2025efficient}
R.~Zhu, E.~Sun, G.~Huang, and O.~Celiktutan, ``Efficient continual adaptation of pretrained robotic policy with online meta-learned adapters,'' {\em arXiv preprint arXiv:2503.18684}, 2025.

\bibitem{xiao2024beyond}
Z.~Xiao and C.~G. Snoek, ``Beyond model adaptation at test time: A survey,'' {\em arXiv preprint arXiv:2411.03687}, 2024.

\bibitem{liu2026fly}
C.~Liu, Y.~Liu, T.~Wang, Q.~Zhuang, J.~C. Liang, W.~Yang, R.~Xu, Q.~Wang, D.~Liu, and C.~Han, ``On-the-fly vla adaptation via test-time reinforcement learning,'' {\em arXiv preprint arXiv:2601.06748}, 2026.

\bibitem{xutest}
S.~Xu, M.~Tan, L.~Liu, Z.~Zhang, P.~Zhao, {\em et~al.}, ``Test-time adapted reinforcement learning with action entropy regularization,'' in {\em Forty-second International Conference on Machine Learning}, 2025.

\bibitem{chentowards}
K.~Chen, H.~Wei, Z.~Deng, and S.~Lin, ``Towards fast safe online reinforcement learning via policy finetuning,'' {\em Transactions on Machine Learning Research}, 2026.

\bibitem{williams1989learning}
R.~J. Williams and D.~Zipser, ``A learning algorithm for continually running fully recurrent neural networks,'' {\em Neural computation}, vol.~1, no.~2, pp.~270--280, 1989.

\bibitem{catfolis1993method}
T.~Catfolis, ``A method for improving the real-time recurrent learning algorithm,'' {\em Neural networks}, vol.~6, no.~6, pp.~807--821, 1993.

\bibitem{bellec2020solution}
G.~Bellec, F.~Scherr, A.~Subramoney, E.~Hajek, D.~Salaj, R.~Legenstein, and W.~Maass, ``A solution to the learning dilemma for recurrent networks of spiking neurons,'' {\em Nature communications}, vol.~11, no.~1, p.~3625, 2020.

\bibitem{rostami2022eprop}
A.~Rostami, B.~Vogginger, Y.~Yan, and C.~G. Mayr, ``E-prop on spinnaker 2: Exploring online learning in spiking rnns on neuromorphic hardware,'' {\em Frontiers in Neuroscience}, vol.~16, p.~1018006, 2022.

\bibitem{10.7554/eLife.43299}
J.~M. Murray, ``Local online learning in recurrent networks with random feedback,'' {\em eLife}, vol.~8, p.~e43299, may 2019.

\bibitem{marschall2020}
O.~Marschall, K.~Cho, and C.~Savin, ``A {{Unified Framework}} of {{Online Learning Algorithms}} for {{Training Recurrent Neural Networks}},'' {\em Journal of Machine Learning Research}, vol.~21, no.~135, pp.~1--34, 2020.

\bibitem{zucchet2023}
N.~Zucchet, R.~Meier, S.~Schug, A.~Mujika, and J.~Sacramento, ``Online learning of long-range dependencies,'' in {\em Advances in Neural Information Processing Systems} (A.~Oh, T.~Naumann, A.~Globerson, K.~Saenko, M.~Hardt, and S.~Levine, eds.), vol.~36, pp.~10477--10493, Curran Associates, Inc., 2023.

\bibitem{gallego2022event}
G.~Gallego, T.~Delbrück, G.~Orchard, C.~Bartolozzi, B.~Taba, A.~Censi, S.~Leutenegger, A.~J. Davison, J.~Conradt, K.~Daniilidis, and D.~Scaramuzza, ``Event-based vision: A survey,'' {\em IEEE Transactions on Pattern Analysis and Machine Intelligence}, vol.~44, no.~1, pp.~154--180, 2022.

\bibitem{maqueda2018event}
A.~I. Maqueda, A.~Loquercio, G.~Gallego, N.~Garc{\'\i}a, and D.~Scaramuzza, ``Event-based vision meets deep learning on steering prediction for self-driving cars,'' in {\em Proceedings of the IEEE conference on computer vision and pattern recognition}, pp.~5419--5427, 2018.

\bibitem{peng2023get}
Y.~Peng, Y.~Zhang, Z.~Xiong, X.~Sun, and F.~Wu, ``Get: Group event transformer for event-based vision,'' in {\em Proceedings of the IEEE/CVF International Conference on Computer Vision}, pp.~6038--6048, 2023.

\bibitem{zheng2023deep}
X.~Zheng, Y.~Liu, Y.~Lu, T.~Hua, T.~Pan, W.~Zhang, D.~Tao, and L.~Wang, ``Deep learning for event-based vision: A comprehensive survey and benchmarks,'' {\em arXiv preprint arXiv:2302.08890}, 2023.

\bibitem{9129849}
G.~Chen, H.~Cao, J.~Conradt, H.~Tang, F.~Rohrbein, and A.~Knoll, ``Event-based neuromorphic vision for autonomous driving: A paradigm shift for bio-inspired visual sensing and perception,'' {\em IEEE Signal Processing Magazine}, vol.~37, no.~4, pp.~34--49, 2020.

\bibitem{9560881}
A.~Vitale, A.~Renner, C.~Nauer, D.~Scaramuzza, and Y.~Sandamirskaya, ``Event-driven vision and control for uavs on a neuromorphic chip,'' in {\em 2021 IEEE International Conference on Robotics and Automation (ICRA)}, pp.~103--109, 2021.

\bibitem{10342437}
S.~N. Aspragkathos, E.~Ntouros, G.~C. Karras, B.~Linares-Barranco, T.~Serrano-Gotarredona, and K.~J. Kyriakopoulos, ``An event-based tracking control framework for multirotor aerial vehicles using a dynamic vision sensor and neuromorphic hardware,'' in {\em 2023 IEEE/RSJ International Conference on Intelligent Robots and Systems (IROS)}, pp.~6349--6355, 2023.

\bibitem{paredes2024fully}
F.~Paredes-Vall{\'e}s, J.~J. Hagenaars, J.~Dupeyroux, S.~Stroobants, Y.~Xu, and G.~C. de~Croon, ``Fully neuromorphic vision and control for autonomous drone flight,'' {\em Science Robotics}, vol.~9, no.~90, p.~eadi0591, 2024.

\bibitem{sutton2018}
R.~S. Sutton and A.~G. Barto, {\em Reinforcement Learning: {{An}} Introduction}.
\newblock {A Bradford Book}, 2018.

\bibitem{murray2019}
J.~M. Murray, ``Local online learning in recurrent networks with random feedback,'' {\em eLife}, vol.~8, p.~e43299, May 2019.

\bibitem{gerstner2014}
W.~Gerstner, W.~M. Kistler, R.~Naud, and L.~Paninski, {\em Neuronal {{Dynamics}}: {{From Single Neurons}} to {{Networks}} and {{Models}} of {{Cognition}}}.
\newblock {Cambridge}: {Cambridge University Press}, 2014.

\bibitem{farsang2024learning}
M.~Farsang, M.~Lechner, D.~Lung, R.~Hasani, D.~Rus, and R.~Grosu, ``Learning with chemical versus electrical synapses does it make a difference?,'' in {\em 2024 IEEE International Conference on Robotics and Automation (ICRA)}, pp.~15106--15112, IEEE, 2024.

\bibitem{farsang2025parallelizationnonlinearstatespacemodels}
M.~Farsang and R.~Grosu, ``Parallelization of non-linear state-space models: Scaling up liquid-resistance liquid-capacitance networks for efficient sequence modeling,'' {\em Advances in Neural Information Processing Systems}, vol.~38, pp.~169137--169163, 2026.

\bibitem{lemmel2025a}
J.~Lemmel, M.~Kranzl, A.~Lamine, P.~Neubauer, R.~Grosu, and S.~Neubauer, ``Online fine-tuning of carbon emission predictions using real-time recurrent learning for state space models,'' {\em arXiv preprint arXiv:2508.00804}, 2025.

\bibitem{resch2025mmdvs}
F.~Resch, M.~Farsang, and R.~Grosu, ``Mmdvs-lf: Multi-modal dynamic vision sensor and eye-tracking dataset for line following,'' {\em arXiv preprint arXiv:2409.18038}, 2024.

\bibitem{walsh17}
C.~Walsh and S.~Karaman, ``Cddt: Fast approximate 2d ray casting for accelerated localization,'' vol.~abs/1705.01167, 2017.

\end{thebibliography}
\bibliographystyle{ieeetr}

%%%%%%%%%%%%%%%%%%%%%%%%%%%%%%%%%%%%%%%%%%%%%%%%%%%%%%%%%%%%

%%%%%%%%%%%%%%%%%%%%%%%%%%%%%%%%%%%%%%%%%%%%%%%%%%%%%%%%%%%%%%%%%%%%%%%%%%%%%%%
%%%%%%%%%%%%%%%%%%%%%%%%%%%%%%%%%%%%%%%%%%%%%%%%%%%%%%%%%%%%%%%%%%%%%%%%%%%%%%%
% APPENDIX
%%%%%%%%%%%%%%%%%%%%%%%%%%%%%%%%%%%%%%%%%%%%%%%%%%%%%%%%%%%%%%%%%%%%%%%%%%%%%%%
%%%%%%%%%%%%%%%%%%%%%%%%%%%%%%%%%%%%%%%%%%%%%%%%%%%%%%%%%%%%%%%%%%%%%%%%%%%%%%%
\newpage
\appendix

\section{RTRRL Algorithm Details}

\paragraph{Hyperparameters}
We first did two rough grid searches over learning rates for the actor and critic, followed by a grid search over entropy regularization strengths. Finally, we ran a large-scale Bayesian hyperparameter tuning run that covered the learning rates and decay schedules for them, as well as the strength of the parameter-change penalty introduced above.

We ran separate grid searches for the actor learning rate, the entropy regularization, the number of environment steps for linear warmup of the actor learning rate (starting from 0), and the parameter change penalty factor. The values tested are summarized in Table~\ref{tab:grid}.

\begin{table}[h]
\centering
\caption{Grid search hyperparameters.}
\label{tab:grid}
\begin{tabular}{ll}
\hline
\textbf{Hyperparameter} & \textbf{Values} \\
\hline
Actor learning rate & $10^{-3}, 10^{-4}, 10^{-5}, 10^{-6}$ \\
Critic learning rate & $10^{-3}, 10^{-4}, 10^{-5}$ \\
Entropy factor & $0, 10^{-3}, 10^{-4}$ \\
Param change penalty factor & $0, 10^{-2}, 10^{-4}, 10^{-5}$ \\
Warmup steps actor LR & $0, 1000, 3000, 10000$ \\
\hline
\end{tabular}

\end{table}

We further ran a Bayesian sweep, testing roughly 100 sets of sampled hyperparameters with ranges given in Table~\ref{tab:bayes}. Values given as intervals were drawn from a log-uniform distribution to accommodate magnitude differences.

\begin{table}[h]
\centering
\caption{Bayesian sweep hyperparameters.}
\label{tab:bayes}
\begin{tabular}{ll}
\hline
\textbf{Hyperparameter} & \textbf{Values} \\
\hline
Actor learning rate & $[10^{-4}, 10^{-7}]$ \\
Critic learning rate & $[10^{-2}, 10^{-5}]$ \\
Param change penalty factor & $0, 10^{-2}, 10^{-5}$ \\
Warmup steps actor LR & $0, 1000, 3000$ \\
\hline
\end{tabular}

\end{table}

We found that a learning rate around $10^{-6}$ for the actor is best. The critic learning rate appeared to be of less importance with values in the range of $10^{-3}$ to $10^{-5}$ being acceptable. Entropy regularization showed a negligible impact overall, so we decided to move forward without it. Finally, the best choice for the parameter change penalty coefficient resulted to be $10^{-5}$. 
\begin{table}[h]
\centering
\caption{RTRRL hyperparameter configuration for fine-tuning.}
\label{tab:hyperparams}
\begin{tabular}{l c}
\hline
Hyperparameter & Value \\
\hline
Discount factor $\gamma$             & $0.99$ \\
Eligibility trace decay factor $\lambda$             & $0.95$ \\
Actor learning rate $\alpha_A$      & \(10^{-6}\) \\
Critic learning rate $\alpha_C$        & \(10^{-5}\) \\
Entropy rate $\eta_H$             & $0$ \\
Parameter change penalty rate $\eta_P$   & \(10^{-5}\) \\
\hline
\end{tabular}
\end{table}

Algorithm~\ref{alg:RTRL} below outlines the steps of RTRRL. In contrast to the original algorithm, we use separate RNNs for the actor and critic.
%\begin{figure}
\begin{algorithm}[ht]
    \caption{Real-Time Recurrent Reinforcement Learning \cite{lemmel2025real}}
    \label{alg:RTRL}
    \begin{algorithmic}[1]
	    \STATE $\theta_{A}, \theta_{C} \gets $ initialize RNN parameters
	    \STATE $h_A, h_C, e_{A}, e_{C} \gets \mathbf 0$ 
		\STATE $o \gets $ initial observation
	    \STATE $\pi, h_A, \hat J_A \gets $RNN$_{\theta_{A}}(o, h_A, \mathbf 0)$ 
        \STATE $\hat v, h_C, \hat J_C \gets $RNN$_{\theta_{C}}(o, h_C, \mathbf 0)$
        % \STATE
		\WHILE{not done}
		    \STATE $a \gets $ sample($\pi$)
		    \STATE $o, r \gets$ execute action $a$
            % \STATE
			\STATE $\pi', h_A, \hat J_A' \gets $RNN$_{\theta_{A}}(o, h_A, \hat J_A)$ 
            \STATE $\hat v', h_C, \hat J_C' \gets $RNN$_{\theta_{C}}(o, h_C, \hat J_C)$ 
           % \STATE
            \STATE $e_{C} \gets \gamma \lambda_{C} e_{C} + \hat J_C \nabla_{h_C} \hat v$
			\STATE $e_{A} \gets \gamma \lambda_{A} e_{A} + \hat J_A \nabla_{h_{A}} \log \pi[a]$

            \STATE $\delta \gets r + \gamma \hat v' - \hat v$ 
			
			\STATE $\theta_{C} \gets \theta_{C} + \alpha_C \delta e_{C}$   
			\STATE $\theta_{A} \gets \theta_{A} + \alpha_{A} \delta e_{A}$

			\STATE $\hat v \gets \hat v', \quad \hat J_A \gets \hat J_A', \quad \hat J_C \gets \hat J_C'$
		\ENDWHILE
    \end{algorithmic}
\end{algorithm}
%\end{figure}

\section{Event-Camera Representation}

%
%TODO: make a different plot out of this
\begin{figure}[H]
    \centering
     \begin{subfigure}[b]{0.4\linewidth}
         \centering
         \includegraphics[width=\textwidth]{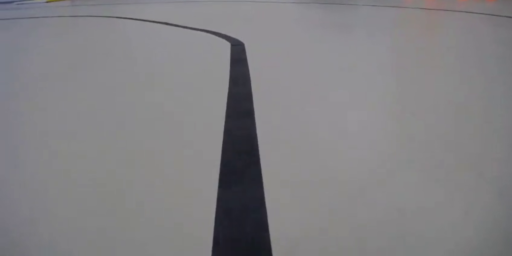}
        \caption{RGB image.}
        \label{fig:rgb}
     \end{subfigure}
     \hfill
     \begin{subfigure}[b]{0.4\linewidth}
         \centering
         {%
            \setlength{\fboxsep}{0pt}%
            \setlength{\fboxrule}{0.33pt}%  
            \fbox{\includegraphics[width=\linewidth]{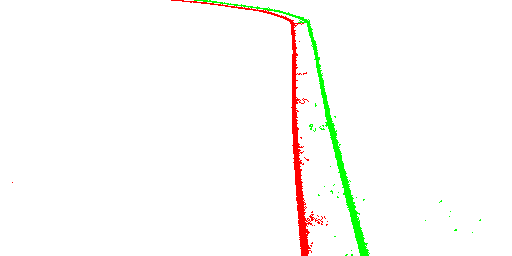}}
        }%
        \caption{Event frame.}
        \label{fig:event-frame}
     \end{subfigure}
     % \hfill
     % \begin{subfigure}[b]{0.245\linewidth}
     %     \centering
     %     {%
     %        \setlength{\fboxsep}{0pt}%
     %        \setlength{\fboxrule}{0.33pt}%  
     %        \fbox{\includegraphics[width=\linewidth]{figures/frame_000700.ts.png}}
     %    }%
     %    \caption{Time surface.}
     %    \label{fig:time-surface}
     % \end{subfigure}
     % \hfill
     % \begin{subfigure}[b]{0.245\linewidth}
     %     \centering
     %     {%
     %        \setlength{\fboxsep}{0pt}%
     %        \setlength{\fboxrule}{0.33pt}%  
     %        \fbox{\includegraphics[width=\linewidth]{figures/frame_000700.et.png}}
     %    }%
     %    \caption{Event tensor.}
     %    \label{fig:event-tensor}
     % \end{subfigure}
    \caption{RGB frame and the corresponding DVS event frame representation.}
    %In the time surface and the event tensor, darker colors indicate earlier events and lighter colors later ones.
    \label{fig:representations}
    % \vspace{-3ex}
\end{figure}

\section{Computing resources}

Pretraining was conducted on NVIDIA L40S GPUs with a runtime of approximately one hour per run. Fine-tuning was done on CPU since it allows faster throughput when working with a batch size of one. One fine-tuning run took around 3 hours on a machine with 64 CPU cores, where up to 4 runs were conducted in parallel.

\section{Additional Pretraining Results}

\begin{figure}[H]
    \centering
    \includegraphics[width=0.45\linewidth]{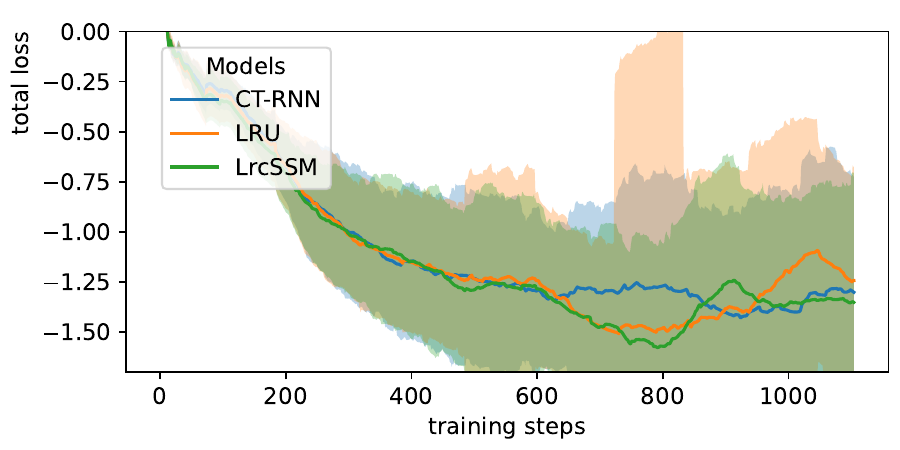}
      \includegraphics[width=0.45\linewidth]{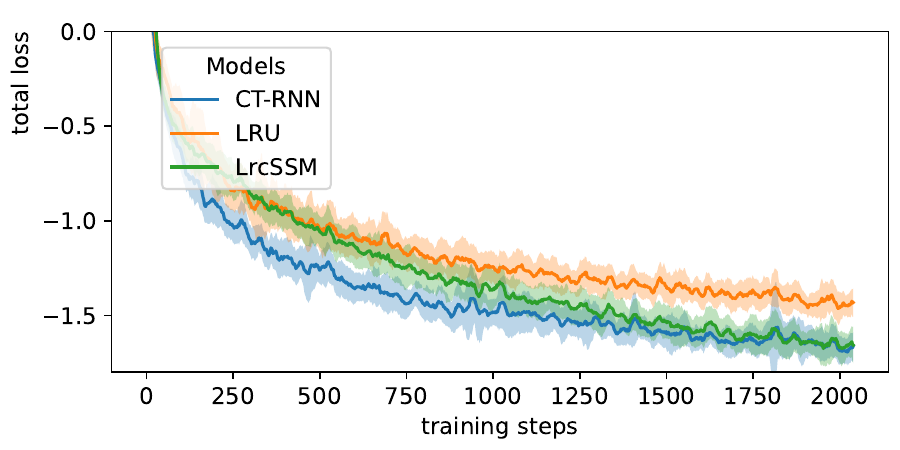}
    \caption{Mean validation loss during pretraining on the \texttt{CarRacing} (left) and \texttt{LineTracking} (right) datasets. Shown is the mean reward of five seeds per model type, with standard deviation shown as shaded regions.}
    \label{fig:carracing_pretraining}
\end{figure}

We show the validation loss curves from pretraining on the \texttt{CarRacing} and \texttt{LineTracking} datasets in Fig.~\ref{fig:carracing_pretraining}.

\begin{figure}[H]
    \centering
    \includegraphics[width=0.6\linewidth]{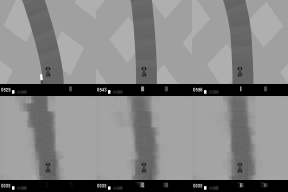}
    \includegraphics[width=0.39\linewidth]{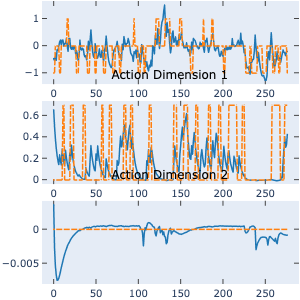}
    
    \caption{Left: Exemplary decoded images predicted by the CNN autoencoder after pretraining on the \texttt{CarRacing} dataset. Right: Actions predicted by the pretrained policy.}
    \label{fig:decoded_carracing}
\end{figure}

\begin{figure}[H]
    \centering
    \includegraphics[width=0.35\linewidth]{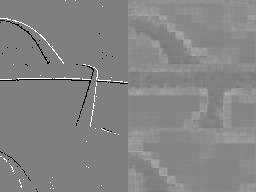}
    \includegraphics[width=0.6\linewidth]{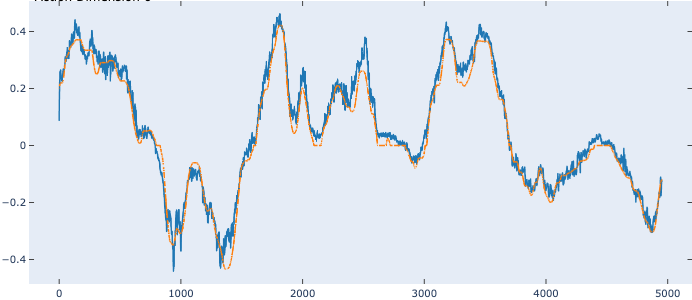}
    \caption{Left: Exemplary decoded images predicted by the CNN autoencoder after pretraining on the \texttt{LineTracking} dataset. Right: Actions predicted by the pretrained policy.}
    \label{fig:decoded_linetracking}
\end{figure}

\section{\textit{LineTracking} Fine-tuning Progress}

\subsection{Pretrained Trajectories}

\begin{figure}[H]
    \centering
     \begin{subfigure}[b]{0.32\linewidth}
        \centering
        \includegraphics[width=\textwidth]{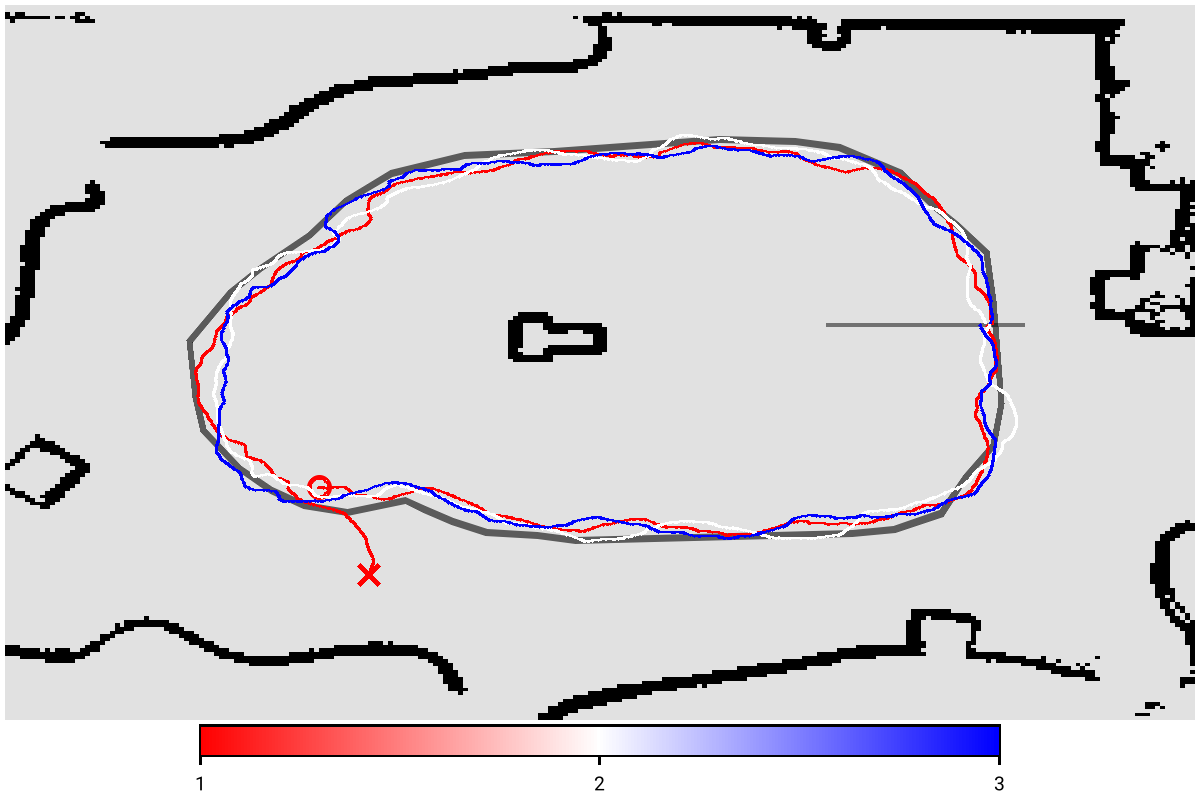}
        \caption{CT-RNN.}
        \label{fig:pre_train_laps_ctrnn}
     \end{subfigure}
     % \hfill
     \begin{subfigure}[b]{0.32\linewidth}
        \centering
        \includegraphics[width=\linewidth]{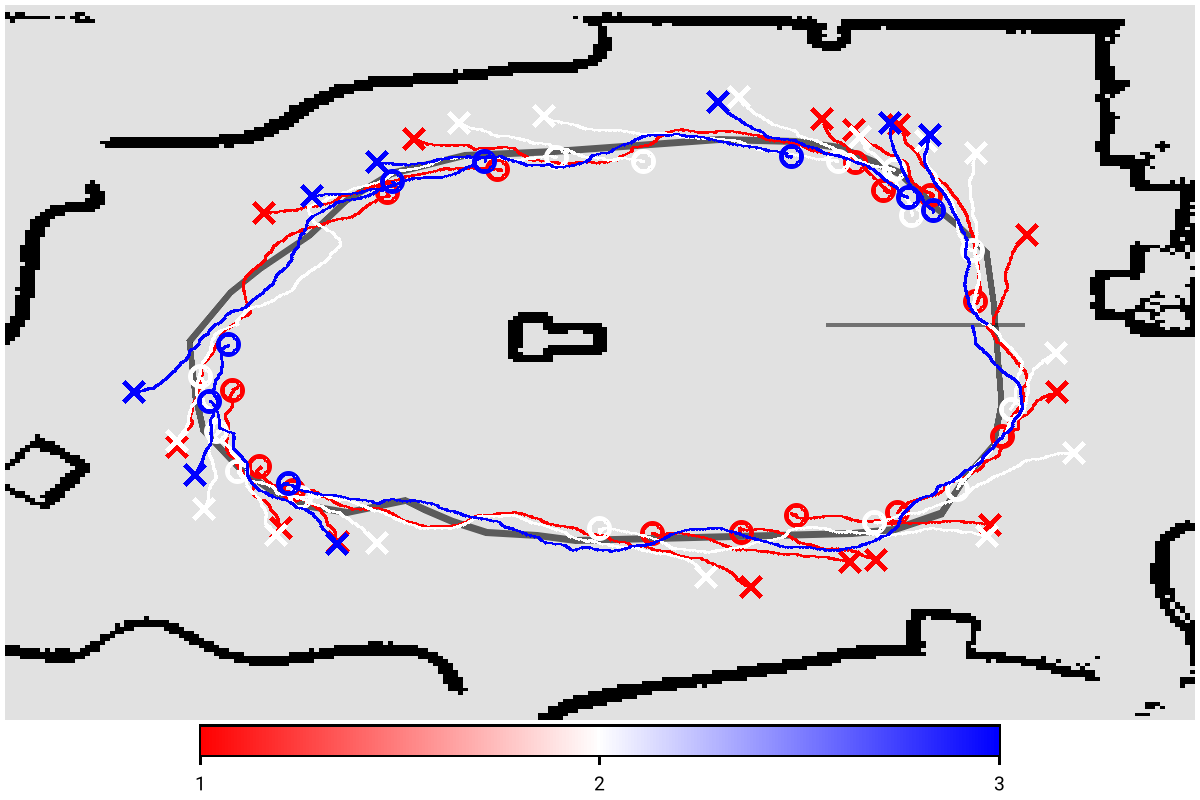}
        \caption{LRU.}
        \label{fig:pre_train_laps_lru}
     \end{subfigure}
     % \hfill
     \begin{subfigure}[b]{0.32\linewidth}
        \centering
        \includegraphics[width=\linewidth]{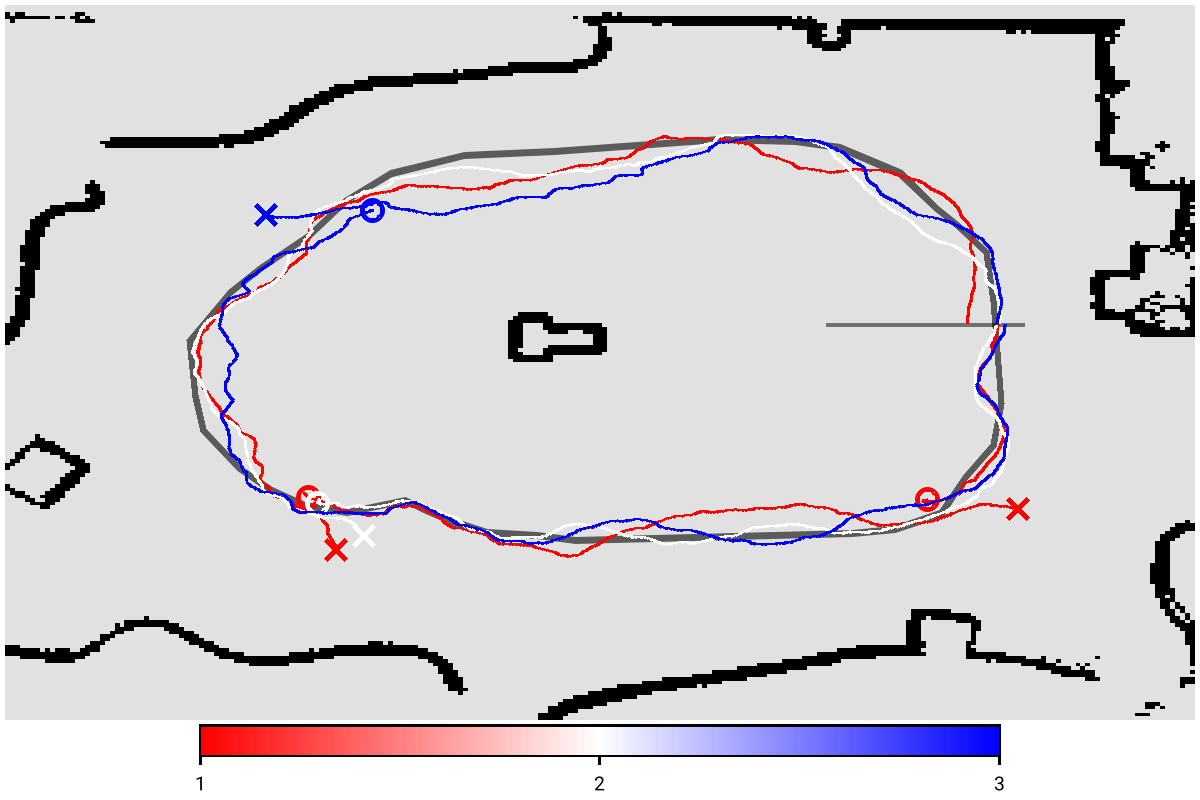}
        \caption{LrcSSM.}
        \label{fig:pre_train_laps_lrc}
    \end{subfigure}
    
    \caption{Trajectories of policies without fine-tuning. Crosses indicate manual intervention, and circles indicate the resumption by the policy.}
    \label{fig:pre_train_laps}
\end{figure}

\begin{figure}[H]
    \begin{subfigure}[b]{\linewidth}
        \centering
        \includegraphics[width=.3\linewidth]{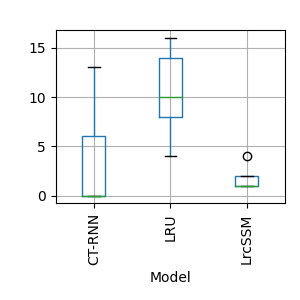}
        \caption{Number of interventions per lap of the pretrained \texttt{LineTracking} models.}
        \label{fig:interventions}
    \end{subfigure}
\end{figure}

\subsection{Policies During Fine-tuning}

\begin{figure}[H]
    \centering
     \begin{subfigure}[b]{0.45\linewidth}
        \centering
        \includegraphics[width=\textwidth]{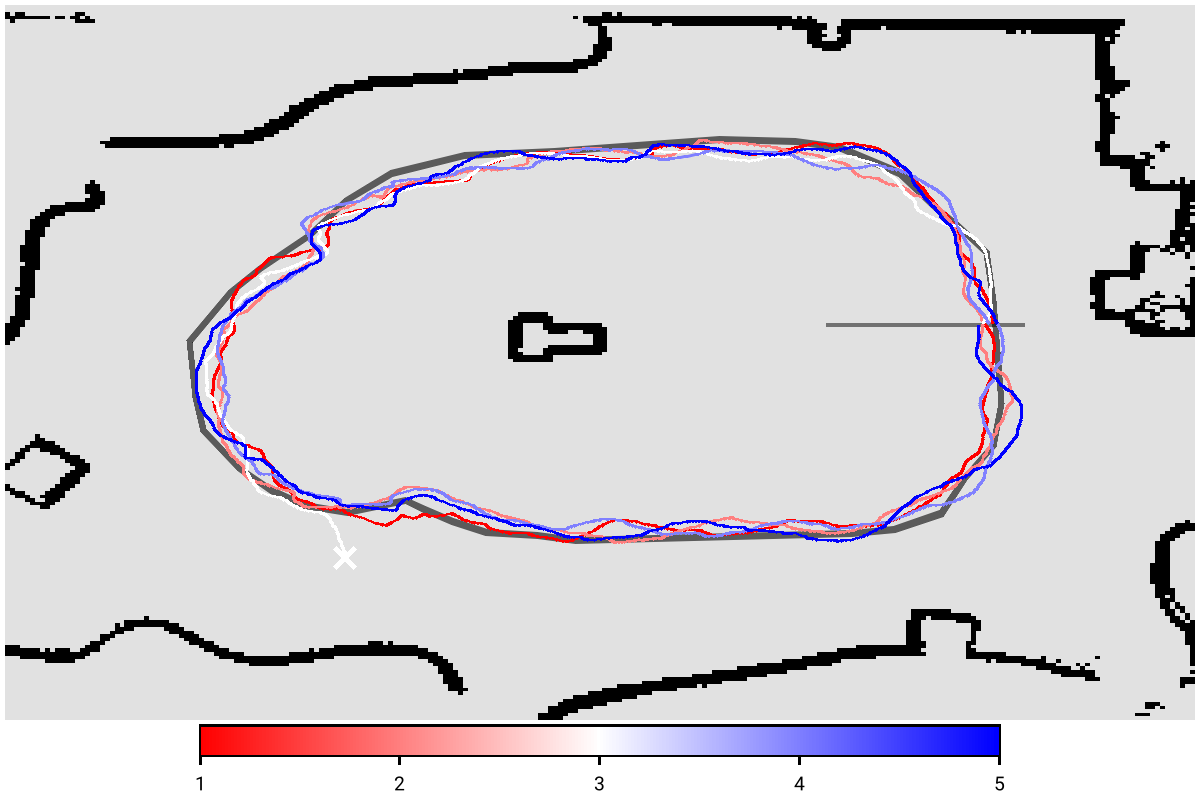}
        \caption{CT-RNN.}
        \label{fig:finetune_laps_ctrnn}
     \end{subfigure}
     \hfill
     \begin{subfigure}[b]{0.45\linewidth}
        \centering
        \includegraphics[width=\linewidth]{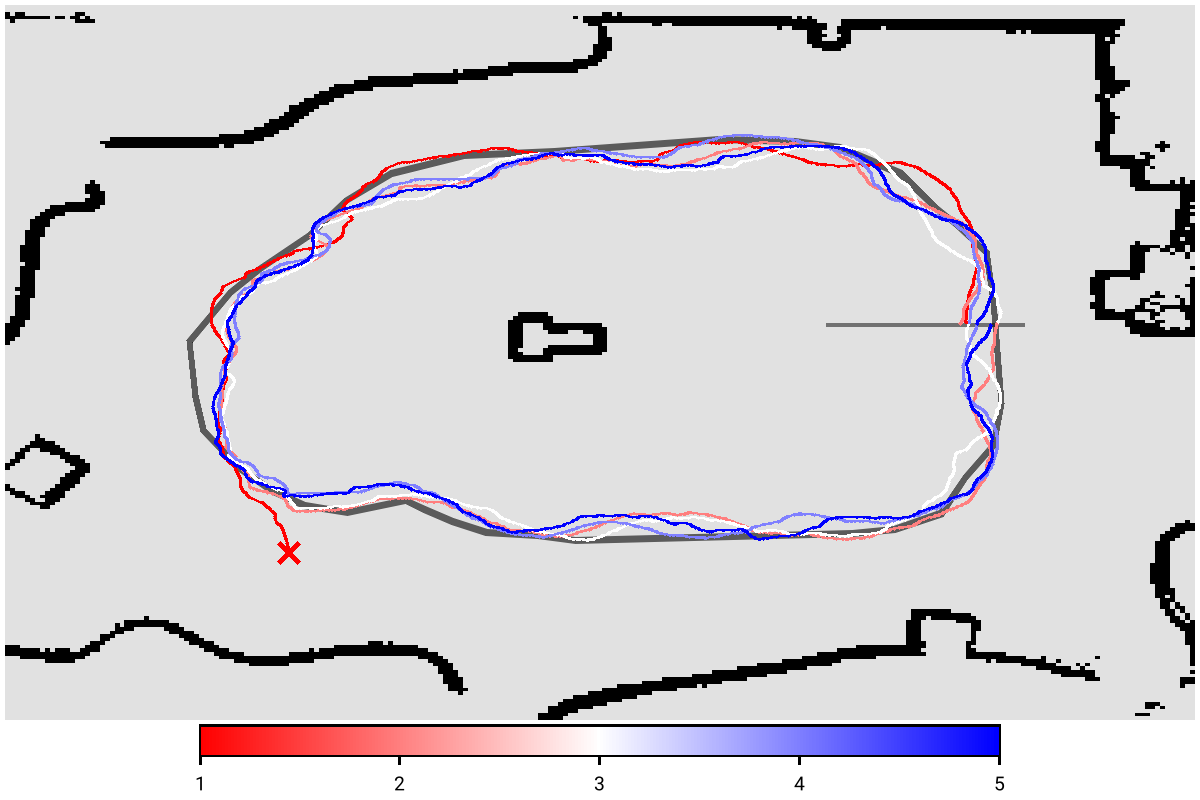}
        \caption{LrcSSM.}
        \label{fig:finetune_laps_lrc}
     \end{subfigure}
    \caption{Trajectories of policies with fine-tuning. Laps that required manual intervention were terminated upon intervention.}
    \label{fig:pre_train_laps2}
\end{figure}

\section{Broader Impact and Limitations}

Although this paper focuses on autonomous driving, we emphasize that online controller adaptation can have unexpected side effects. Neural network behavior is hard to verify, and this difficulty is exacerbated when weights are changed during operation. We therefore strongly discourage the use of the presented method in any safety-critical applications, such as steering vehicles carrying humans. The real-world experiments presented in this paper merely serve as a proof-of-concept, and a more thorough evaluation is needed before applying the method – this is also a limitation of this work.

That being said, we would like to highlight that a bad choice of hyperparameters led to passive degradation of the policy. More precisely, the predicted action distribution's entropy continues to increase as long as the temporal-difference error is negative, which in turn results in a broader range of actions being taken. For the \texttt{CarRacing} environment in particular, this leads to more breaking and a reduction in speed, culminating in a stationary car.

 Another limitation of our work is the lack of baselines for comparison. RTRRL is fully online with updates computed at each step, setting it apart from all state-of-the-art RL algorithms. A direct comparison to replay-based methods is therefore not possible. Nonetheless, we present results for PPO-based fine-tuning in the next section.

\section{PPO Baseline}

To compare the proposed method with a state-of-the-art reinforcement learning algorithm, we ran fine-tuning experiments using PPO on the \texttt{CarRacing} environment. We used the same pretrained models, fine-tuned via RTRRL, and the same hyperparameters. Table \ref{tab:ppo} shows the results in comparison to the RTRRL results. PPO achieves similar results but requires a replay buffer and separate training phases, while RTRRL applies updates in each step.

\begin{table}[h]
\centering
\caption{PPO fine-tuning results on the CarRacing environment in comparison to RTRRL.}
\label{tab:ppo}
\begin{tabular}{lc}
\toprule
Model  & Evaluation reward \\
\midrule
LrcSSM & 775.35 $\pm$ 58.93 \\
LrcSSM PPO & 523.21 $\pm$ 146.94 \\
LRU & 787.76 $\pm$ 58.90 \\
LRU PPO & 826.67 $\pm$ 48.50 \\
CT-RNN & 747.79 $\pm$ 26.61 \\
CT-RNN PPO & -104.43 $\pm$ 280.85 \\
\bottomrule
\end{tabular}

\end{table}

%%%%%%%%%%%%%%%%%%%%%%%%%%%%%%%%%%%%%%%%%%%%%%%%%%%%%%%%%%%%

% \newpage
% \input{checklist.tex}

\end{document}